\documentclass{article}

\usepackage{microtype}
\usepackage{graphicx}
\usepackage{subcaption}
\usepackage{booktabs}
\usepackage{multirow}
\usepackage{float}  %
\usepackage{hyperref}

\usepackage[preprint]{icml2027}
\makeatletter
\renewcommand{\printAffiliationsAndNotice}[1]{\global\icml@noticeprintedtrue}
\makeatother
\hypersetup{pdftitle={How to Turn an Attention Neural Network into an Interpretable Hard-Logic Predicate Circuit: Truth-Table Predicate Extraction and State-Machine Merging},
  pdfauthor={Yuchen Li, Zongqi Fan, Xintong Li, Nguyen H. Tran, Ken-Tye Yong}, pdfsubject={}}

\usepackage{amsmath}
\usepackage{amssymb}
\usepackage{mathtools}
\usepackage{amsthm}
\usepackage[capitalize,noabbrev]{cleveref}
\usepackage{tikz}
\usetikzlibrary{arrows.meta,positioning,fit,calc}
\usepackage{pgfplots}
\pgfplotsset{compat=1.18}
\usepgfplotslibrary{groupplots}
\definecolor{seriesorder}{HTML}{2A78D6}
\definecolor{seriesrule}{HTML}{EB6834}

\theoremstyle{plain}
\newtheorem{theorem}{Theorem}[section]
\newtheorem{proposition}[theorem]{Proposition}

\theoremstyle{definition}
\newtheorem{definition}[theorem]{Definition}
\theoremstyle{remark}

\newcommand{\sd}[1]{{\scriptsize\,$\pm$\,#1}}  %
\renewcommand{\algorithmiccomment}[1]{\hfill{\color{black!60}$\triangleright$~#1}}

\newcommand{\cls}{\texttt{[CLS]}}
\newcommand{\Sel}{\mathrm{Sel}}
\newcommand{\ON}{\mathrm{ON}}
\newcommand{\OFF}{\mathrm{OFF}}

\newcommand{\agr}{\operatorname{agr}}
\newcommand{\eqR}[1]{\mathrel{\overset{#1}{=}}}
\newcommand{\neqR}[1]{\mathrel{\overset{#1}{\neq}}}

\newcommand{\orderSel}{order selection}
\newcommand{\ruleSel}{rule selection}
\newcommand{\OrderSel}{Order selection}
\newcommand{\RuleSel}{Rule selection}

\icmltitlerunning{Attention Networks as Interpretable Hard-Logic Predicate Circuits}

\begin{document}

\twocolumn[
  \icmltitle{How to Turn an Attention Neural Network into an Interpretable\\
    Hard-Logic Predicate Circuit: Truth-Table Predicate Extraction\\
    and State-Machine Merging}

  {\begin{icmlauthorlist}
    {\bf Yuchen Li} \quad {\bf Zongqi Fan}$^{*}$ \quad {\bf Xintong Li}$^{*}$ \quad {\bf Nguyen H. Tran}$^{\dagger}$ \quad {\bf Ken-Tye Yong}$^{\dagger}$\\[4pt]
    {\small The University of Sydney}\\[2pt]
    {\small $^{*}$Joint second authors. \quad $^{\dagger}$Corresponding authors.}\\[4pt]
    {\small Preliminary version, November 2026}
  \end{icmlauthorlist}}

  \icmlkeywords{logic minimization, neural network to circuit conversion, Transformers, attention, interpretability, FPGA, hardware efficiency}

  \vskip 0.3in
]

\printAffiliationsAndNotice{}

\begin{abstract}
We convert a trained attention network as a whole, attention included, into a hard-logic predicate circuit that runs in its place. Every unit of the circuit is a 0/1 predicate on tokens, and every map between predicates is a minimised truth table. The interfaces between blocks are defined from the output down by what the circuit above reads, every block is fitted once on the teacher's own states, and the rules of the complete chain are then refitted on the states that the chain itself produces. Attention selection takes two forms, a row-wise comparison of integer scores (\orderSel) and a learned rule (\ruleSel). Both are fitted on the ordinary training set alone, and only on the examples that the teacher classifies correctly, so that the circuit imitates the teacher where the teacher is right. On TweetEval-emotion, the complete circuit of DistilGPT2 reaches a test gold accuracy of 67.7\,$\pm$\,0.8\% with \orderSel\ and 56.4\,$\pm$\,1.4\% with \ruleSel\ over five seeds, against 74.7\% for the teacher and 64.0\% for a bag of word pieces fitted on the same tweets, and that of BERT-base truncated to 6 blocks reaches 71.1\,$\pm$\,1.1\% and 61.8\,$\pm$\,5.0\% against 80.0\% (bag of words 63.8\%). On four GLUE tasks, \orderSel\ reaches a mean test gold accuracy of 70.6\% (DistilGPT2) and 70.9\% (BERT), above the bag of words (68.6\% and 68.4\%). The circuit needs no multiplier and no weight traffic, and its exported hardware description reproduces it bit for bit. By a synthesis-based estimate averaged over the five tasks, it classifies one example in 1.1--2.0\,$\mu$s on an FPGA card, 180--420 times faster than the network on a GPU at batch 1 with CUDA graphs (measured, 0.35--0.46\,ms), and in 3.3--6.4\,$\mu$s with 25--96\,$\mu$J on a palm-sized board of 7.5--15\,W, 550--2,400 times less energy per example than the GPU, which suits edge devices and miniaturised systems.
\end{abstract}

\section{Introduction}
\label{sec:intro}

A trained neural network computes with soft weights. A single weight means nothing on its own, and what the network does is spread over millions of them, so the network cannot be read from its weights. It can be read from how each layer transforms its features. We record this transformation, layer by layer, as a map between \emph{predicates}, 0/1 features with a definite extension (the tokens on which they hold) and a definite role (the rules above that read them), joined only by AND, OR, NOT and implication. Each map is fitted to be equivalent to its layer on the features that the data reach, so that it keeps the computation the data use, and the maps compose into a circuit that runs in place of the network. Interpretation then means reading predicates and rules instead of weights. A bag of words is interpretable too, but only as additive word weights. It cannot express logical combinations such as the conjunction or the negation of words, nor the relations between words that attention establishes, and the circuit's rules state exactly these.

Why should a small circuit be enough? The network has a huge parameter space, but the data and the internal representations of a trained network occupy low-dimensional manifolds of its high-dimensional spaces \citep{ansuini2019intrinsic}, and fine-tuning a pretrained language model acts in a space of low intrinsic dimension \citep{aghajanyan2021intrinsic}. Low-rank adaptation builds on this: a low-rank update of the pretrained weights is enough to adapt the model to a task \citep{hu2022lora}. The update, however, runs on top of the whole pretrained network, and none of its parameters can be removed at inference. The part of the network that the data of a task use is small, and a small predicate circuit can describe it in place of the whole network. The circuit only has to match the network on the range of features that the data reach, input combinations outside that range do not constrain it, units that switch together on the data form one predicate, and states that the layer above treats alike form one state. This gives the method: we observe the input and output features of each layer on the training data and fit a transformation between predicates to it, a minimised truth table or a set of implication rules.

Such a circuit also implements the trained network directly: inference needs no multiplier and no floating-point arithmetic, and the only stored object is a read-only table from input tokens to their first predicates, so no weight travels between memory and a processor.

Three obstacles stand in the way. The first is attention. Existing conversions tabulate functions whose second operand is a fixed weight \citep{nazemi2018nullanet,umuroglu2020logicnets,tang2023lutnn,blalock2021multiplying}, while attention multiplies two activations. LEAF \citep{cox2026leaf} names this obstacle and keeps such products in arithmetic form. The second obstacle is depth. A circuit fitted to one layer can be accurate on its own, but stacked circuits feed each other inputs they were never fitted on, and their errors compound \citep{ross2011reduction}. The third obstacle is the teacher. A small circuit cannot hold everything that a large network computes, and fitting it to every decision of the teacher spends part of its few distinctions on the teacher's own errors, which can be frequent on a small training set (\cref{sec:fitset}).

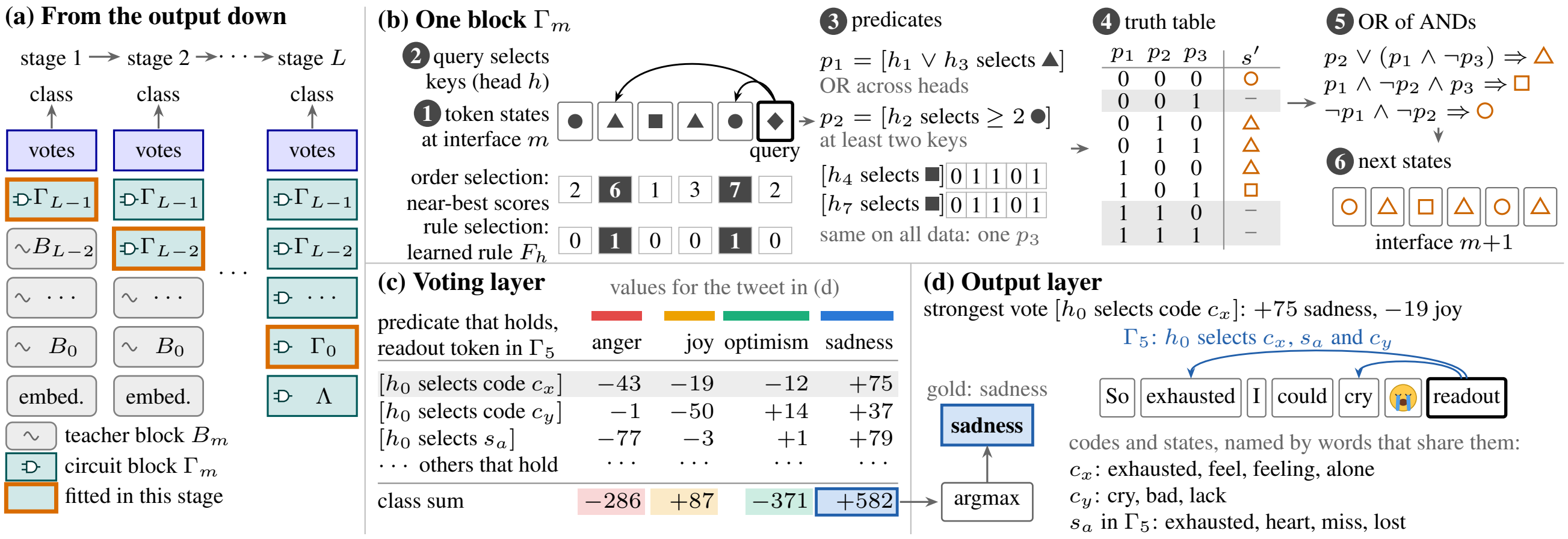
We remove these obstacles with one framework (\cref{fig:overview}). Its principle is that every map of the circuit is a minimised truth table between predicates. Once query and key are short codes, their product is a small table that is minimised exactly. Counting the selected keys is a table with a known minimal circuit, and transitions are tables observed on data, minimised with tolerance to noise into implication rules. The selection of keys takes one of two forms. With \orderSel\ it compares integer scores within a row, and with \ruleSel\ it is learned as a rule like the transitions. The interfaces are defined from the output down, by what the circuit above reads, each block is fitted on the teacher's own states, and the rules of the complete chain are refitted once on the states that the chain itself produces. Every part is fitted on the ordinary training set of the task alone, and only on the training examples that the teacher classifies correctly. The circuit thus imitates the teacher where the teacher is right, and on the other examples it answers from what it learned elsewhere. Our contributions are the following.
\begin{itemize}
  \item \textbf{Method: feature-to-feature truth-value predicatisation of a whole attention network.} To our knowledge, this is the first post-hoc conversion of every block of a pretrained attention network, attention included, into a hard-logic predicate circuit that runs in its place. Every map becomes a minimised truth table or a set of implication rules, minimised exactly, in closed form or with tolerance to noise according to the width of its input, and the query--key product, a product of two activations, becomes a minimised table, so that attention itself becomes logic.
  \item \textbf{Theory: errors add rather than compound.} Interfaces defined from the output down by what the block above reads bound the error of the circuit by the sum of the contract errors of its blocks (Proposition~\ref{prop:additive}), and shorter rules come with a tighter guarantee on unseen tokens (Proposition~\ref{prop:occam}).
  \item \textbf{Accuracy on five tasks and two architectures.} On a decoder (DistilGPT2) and an encoder (BERT-base) over five text-classification tasks, the circuit is above a bag of words fitted on the same examples on average and comes within 1.5 points of BiBERT, the closest earlier work, with logic alone and 2.6 times fewer bits. The conversion can stop at any block, and with the top block converted the network keeps 92--93\% agreement with its teacher.
  \item \textbf{Interpretability: readable by construction.} The relations between predicates are the circuit itself: every rule states which predicates of a token and of its selected keys send it to which state, and every vote states how much a predicate speaks for a class. A reading of the circuit is exact for every decision it makes, unlike a surrogate fitted beside the network, and because each block is fitted on the teacher's own states, it gives a block-by-block account of the teacher.
  \item \textbf{Hardware: microseconds without multipliers.} To our knowledge, this is the first bit-exact hardware description of a complete pretrained attention network converted into logic, without multipliers or weight traffic. By a synthesis-based estimate over five tasks, it classifies an example in 1.1--2.0\,$\mu$s on an Alveo U55C card, 180--420 times faster than the network on a GPU and, for BERT at the same clock, 1.7 times faster than an idealised bound of BiBERT on the same card, and in 3.3--6.4\,$\mu$s on a palm-sized Kria KV260 board with 550--2,400 times less energy per example than the GPU.
\end{itemize}

\section{Related Work}
\label{sec:related}

\paragraph{Logic and lookup-table networks.}
In LogicNets \citep{umuroglu2020logicnets}, sparse low-bit networks are trained so that every neuron is a truth table of small fan-in, NullaNet \citep{nazemi2018nullanet,nazemi2021nullanettiny} realises the neurons of feedforward networks as minimised Boolean functions, and differentiable logic gate networks \citep{petersen2022deep} learn gates directly. Among Transformers, LL-ViT \citep{nag2025llvit} learns lookup-table neurons for the channel mixer and keeps a self-attention token mixer, gate-level Boolean attention networks \citep{shi2025gate} are trained as logic from scratch, and LEAF \citep{cox2026leaf} tabulates the products of attention with constant weights and leaves the products of two activations in arithmetic form. LUT-NN \citep{tang2023lutnn} and MADDNESS \citep{blalock2021multiplying} replace products with a known weight matrix by table lookups. These networks are trained as logic, or they tabulate products with a fixed weight. We convert a given pretrained network after training, and the products of two activations become small tables as well.

\paragraph{Programs and Transformers.}
RASP and Tracr \citep{weiss2021thinking,lindner2023tracr} describe programs and compile them into Transformer weights, and Transformer programs \citep{friedman2023learning,laidang2025adaptive} train constrained Transformers that can be read back as discrete programs. Both go from a program to a network, or train a network to be a program. We go the other way, from a standard pretrained network to a circuit.

\paragraph{Rules from networks.}
TREPAN \citep{craven1995extracting} extracts a symbolic surrogate by querying a trained network, and DeepRED \citep{zilke2016deepred} describes each layer of a small feedforward network by rules over the layer below. Such layer-wise extraction often stops at memory or time limits, or falls to chance level on larger tasks \citep{zilke2016deepred,shams2021rem,espinosazarlenga2021eclaire}. Rule learners such as FOIL \citep{quinlan1990learning} and RIPPER \citep{cohen1995fast}, and rule sets trained by gradient methods \citep{wang2021scalable,qiao2021learning}, learn rules from data. Such rules explain a network as a surrogate of its outputs, or they come from small networks. To our knowledge, no earlier work converts a pretrained Transformer into logic after training and reports a text-classification accuracy. Our rules are fitted block by block inside a pretrained attention network, and the composed circuit replaces the network.

\paragraph{Progressive replacement, low precision and minimisation.}
BERT-of-Theseus \citep{xu2020bert} compresses BERT by replacing its modules at random with smaller soft modules, with a replacement probability that grows during training. Its substitutes stay soft and are trained, whereas our conversion fits each block of hard logic once, on the teacher's own states, and leaves no soft part. Incremental Network Quantization \citep{zhou2017incremental} quantises a network in parts, and BinaryBERT \citep{bai2021binarybert} and BiBERT \citep{qin2022bibert} binarise BERT. Fully binarised by the BinaryBERT recipe, or without a full-precision teacher, BERT-base falls to the majority class on several GLUE tasks \citep{qin2022bibert,xing2024bipft}. Dedicated binarisation schemes recover the accuracy \citep{qin2022bibert,liu2022bit}, but they remain arithmetic networks retrained end to end, whereas our circuit is converted after training and consists of hard logic. Quine--McCluskey enumerates the prime implicants of a Boolean function \citep{quine1952problem,mccluskey1956minimization}, Espresso expands and covers heuristically \citep{brayton1984logic,rudell1987multiple}, multilevel synthesis exploits satisfiability and observability don't-cares \citep{brayton1990multilevel}, and states of a sequential circuit that no input distinguishes can be merged \citep{huffman1954synthesis}. We use these notions between the layers of a network.

\begin{figure*}[t]\centering \input{figures/method_panels} \input{figures/method_panels_caption} \label{fig:overview}\end{figure*}

\section{Method}
\label{sec:method}

\subsection{Setting}
\label{sec:setting}

The teacher $f$ maps a token sequence $x=(x_0,\dots,x_{n-1})$ to one of $C$ classes. It consists of an embedding, blocks $B_0,\dots,B_{L-1}$ with $H$ heads each and a classifier on a \emph{readout token}, \cls\ in an encoder and the final token in a decoder. \emph{Interface} $m$ is the input of block $B_m$. At every interface each token carries a discrete \emph{state} $s_i$, and every token carries a static \emph{lexical code} $c_i$ that depends only on its input token. A \emph{predicate} is a 0/1 function of these symbols, such as $[s_i=x]$, and a \emph{literal} is a predicate or its negation. A \emph{circuit} is built from AND, OR and NOT gates, small integer adders and comparators, and we call such a circuit \emph{hard logic}. The \emph{agreement} of two classifiers $g$ and $g'$ on a set $\mathcal{D}$ is $\agr(g,g')=\Pr_{x\in\mathcal{D}}[g(x)=g'(x)]$. The teacher is the fine-tuned network quantised to FP8 after training (\cref{sec:setup}).

\subsection{Imitating the teacher where it is right}
\label{sec:fitset}

A circuit fitted to all decisions of its teacher, as in distillation \citep{hinton2015distilling}, learns the teacher's errors along with its knowledge, and these errors are not rare: the DistilGPT2 teacher errs on 6.9\% of its training tweets (\cref{app:fitset}). A small circuit that copies these errors spends some of its few distinctions on decisions that do not carry over to new examples. We therefore fit every part of the circuit only on the training examples that the teacher classifies correctly (\cref{app:protocol} lists their numbers). On these examples the teacher's label equals the gold label, so the circuit imitates the teacher only where the teacher is right.

\subsection{One principle: every map is a minimised truth table}
\label{sec:principle}

\paragraph{Intuition.} Each layer is described by what it does to its features. We run the network on the training data, record the input and output features of each layer, and fit a transformation between predicates in its place. A trained network does not use its features independently, and on the data many of its features switch on and off together. A group of predicates that always switch together carries a single bit, so it is treated as one predicate, just as identical columns of a truth table can be merged. What remains is a truth table over few predicates, observed only on the rows that the data reach, and every other row is a don't-care. Such a table is written with three connectives: a conjunction for conditions that must hold together, a negation for a condition that must be absent, and a disjunction for the alternative ways to reach the same output. Every map is therefore an OR of ANDs, and minimisation looks for the shortest such expression that agrees with the observed rows. The same principle serves every map of the network, from the code maps to the attention selection and the state transitions, and only the way of minimising changes with the width of the table, from exact to tolerant of noise.

\begin{definition}[Observed truth table]
Let a map send input codes $u\in\{0,1\}^a$ to output values $y$, and let a sample $\mathcal{S}$ be given. The \emph{care set} $\mathcal{C}=\{u(x):x\in\mathcal{S}\}$ holds the input codes that occur. For each output bit $\beta$ the ON set is $\ON_\beta=\{u\in\mathcal{C}:y_\beta(u)=1\}$ and the OFF set is $\OFF_\beta=\mathcal{C}\setminus\ON_\beta$. Every code outside $\mathcal{C}$ is a \emph{satisfiability don't-care}.
\end{definition}

A \emph{cube} is a conjunction of literals, and a \emph{cover} of $\ON_\beta$ is a set of cubes that every point of $\ON_\beta$ satisfies and no point of $\OFF_\beta$ satisfies. Cubes may extend into the don't-cares. Output distinctions that no downstream gate reads are free as well, as observability don't-cares \citep{brayton1990multilevel}, so a map produces only what the circuit above reads (\cref{sec:interfaces}).

\paragraph{(a) Short fixed-width inputs: exact minimisation.}
A code map from a state or a lexical code to a short query or key code, and an integer score table over a pair of codes, are complete tables of a few hundred entries per head. Every entry is a care point, and only bit patterns that encode no state or code are don't-cares. Each output bit is minimised exactly, by enumerating its prime implicants \citep{quine1952problem,mccluskey1956minimization} and selecting a minimum cover, so the result is exact on every entry.

\paragraph{(b) Row-wide symmetric functions: closed-form minimal circuits.}
Counting the selected keys of a state or code reads a whole row of keys. This function is fixed in advance and symmetric in the keys, and its minimal circuit, a small saturating counter, is known, as is that of the row-wise comparison of \orderSel\ (\cref{sec:selection}). Both are written down directly and are correct on every input.

\paragraph{(c) Feature-to-feature maps: noise-tolerant minimisation.}
The next state of a token depends on its own state and on many count predicates. Its table is wide, it is observed only on data, and its targets carry the noise of the soft network. We minimise it with tolerance to this noise into implication rules (\cref{sec:rules}), and \ruleSel\ learns the selection of each head in the same way. Without noise, the procedure reduces to the exact minimisation of the observed table.

\subsection{Interfaces defined by use}
\label{sec:interfaces}

The states of every interface are defined from the top down, by what the circuit above must distinguish, starting from the states of the readout token that separate the teacher's classes. The states of interface $m$ are found on the teacher, supervised by those of interface $m+1$. States that the circuit above, run on the data, hardly tells apart are merged, as indistinguishable states are merged in sequential circuits \citep{huffman1954synthesis}, and predicates that carry the same information are merged likewise. Each layer thus learns only the distinctions that the layers above use, and every distinction it drops is an observability don't-care.

\subsection{Attention as short code tables}
\label{sec:attention}

Attention multiplies a query by a key, two activations. The circuit maps the query of each head to a short code $Q_h(s_i)$ of the token's state, and the key to short codes of the key's state and of its lexical code. The score $\sigma_h(i,j)$ of a pair is then a sum of small integer tables over pairs of codes and over the position of the key. A product of two activations thus becomes a table of a few hundred entries, which regime (a) minimises exactly. The circuit replaces softmax by a 0/1 selection $\Sel_h(i,j)$ and reads only this relation, through count predicates over the states and codes of the selected keys (regime b). In a decoder, a token selects only among the keys at or before it. The count predicates and the literals of the token's own state form the \emph{predicate bank} of the block.

\subsection{Two forms of selection}
\label{sec:selection}

\paragraph{\OrderSel.} The circuit keeps every key whose integer score is close to the best one in its row,
\begin{equation}
  \Sel_h(i,j)=\bigl[\sigma_h(i,j)\ge\max\nolimits_{k}\sigma_h(i,k)-\delta\bigr],
  \label{eq:sel}
\end{equation}
with a small integer margin $\delta$, a row-wise order predicate whose closed-form circuit is a running maximum over the row and one comparison per key.

\paragraph{\RuleSel.} The selection is itself learned as an implication rule, like the transitions. Its target is the teacher's own attention at the layer, made hard, and its inputs are the codes of query and key and the position of the key relative to the query and to the ends of the tweet. The rule of each head is compiled into a one-bit table over these inputs, which the circuit reads,
\begin{equation}
  \Sel_h(i,j)=F_h\bigl(Q_h(s_i),K_h(s_j,c_j),\mathrm{pos}(i,j)\bigr).
  \label{eq:selrule}
\end{equation}
The score tables are then not part of the circuit, and every step before the readout is a rule.

\subsection{Transitions as rules, the readout as integer votes}
\label{sec:rules}

Each token moves to a state of the next interface by implication rules over the predicate bank of its block, \emph{if these predicates hold, then the token moves to this state}. The rules come from the noise-tolerant minimisation of regime (c), which keeps only the conjunctions of predicates that the data support beyond the noise of the soft targets and prefers short rules. Every token receives exactly one next state.

At the last block, the predicate bank $z\in\{0,1\}^n$ of the readout token decides the class by integer votes,
\begin{equation}
  \hat{y}=\arg\max\nolimits_c\Bigl(b_c+\textstyle\sum_{\ell:\,z_\ell=1}W_{\ell c}\Bigr),
  \label{eq:votes}
\end{equation}
with small integers $W$ and $b$. Because $z$ is 0/1, the sum only adds the votes of the predicates that hold. The readout is a threshold function of the predicates, and its circuit is an adder tree and a comparator. Sorting the votes of each class gives a scorecard of the predicates that speak for and against it.

\subsection{Conversion from the output down}
\label{sec:conversion}

\begin{algorithm}[t]
  \caption{Conversion from the output down (overview, every setting in \cref{app:impl})}
  \label{alg:conversion}
  \begin{algorithmic}[1]
    \REQUIRE teacher $f$ with blocks $B_0,\dots,B_{L-1}$, training tweets $X$ with gold labels $y$
    \STATE $X^{+}\gets\{x\in X:f(x)=y(x)\}$ \COMMENT{\cref{sec:fitset}}
    \FOR{$m=L-1$ \textbf{down to} $0$}
      \STATE define the states of interface $m$ on the teacher by their use above \COMMENT{\cref{sec:interfaces}}
      \STATE $\Gamma_m\gets$ block $B_m$ as code tables, selection and rules, fitted on the teacher's states
      \STATE fit the votes $\rho$ of the hybrid of $B_0,\dots,B_{m-1}$ and $\Gamma_m,\dots,\Gamma_{L-1}$
    \ENDFOR
    \STATE refit the rules and votes of $\Gamma_0,\dots,\Gamma_{L-1},\rho$ on the states that the chain produces
    \STATE $\Lambda\gets$ table from (token, position) to the states and codes of interface 0
    \STATE \textbf{return} $(\Lambda,\Gamma_0,\dots,\Gamma_{L-1},\rho)$
  \end{algorithmic}
\end{algorithm}

\Cref{alg:conversion} converts the network from the output down (\cref{fig:overview}a). Every fit in it uses only the teacher-correct training tweets $X^{+}$. Each block is fitted on the teacher's own states at its two interfaces, so that the teacher stays unchanged throughout. After the last block, the rules and votes of every circuit block are refitted on the states that the complete chain itself produces, with the states each block was built to produce as targets, so that every block learns from the inputs it receives at inference. The deployed circuit computes $s^0=\Lambda(x)$, then $s^{m+1}=\Gamma_m(s^m)$ for $m=0,\dots,L-2$, and finally the class $\rho(z)$ from the bank $z$ of the readout token in $\Gamma_{L-1}$.

\paragraph{Why from the output down.} When block $m$ is converted, the circuit above it is fixed, so what it reads, and with it the don't-cares and merges of interface $m+1$, is explicit. After every step, the teacher's lower blocks followed by the circuit blocks form a complete classifier (\cref{tab:stages}).

\subsection{Theory}
\label{sec:theory}

Because every interface is defined by what the block above reads, the errors of the blocks add rather than compound: the error of the complete circuit is at most the sum of the contract errors of its blocks and its readout (Proposition~\ref{prop:additive} in \cref{app:theory}). The contract of a block is defined on the teacher's states at its two interfaces, the very states on which the conversion fits it. A second, standard result (Proposition~\ref{prop:occam}) bounds the error of a rule set on unseen tokens by its training error and its description length, so that shorter rules come with a tighter guarantee.

\section{Experiments}
\label{sec:experiments}

\subsection{Setup}
\label{sec:setup}

\paragraph{Data and models.} The main task is TweetEval-emotion \citep{barbieri2020tweeteval}, four classes with its official splits, on which we also analyse the circuit. Four tasks of GLUE \citep{wang2019glue} extend the conversion to other text and to sentence pairs: SST-2 \citep{socher2013recursive}, QQP, MRPC \citep{dolan2005automatically} and CoLA \citep{warstadt2019neural}. The GLUE validation split serves as test (for QQP a fixed part of it), since the GLUE test labels are not public, a stratified part of the GLUE training split serves as validation, and a sentence pair enters as one sequence with the separator token of each network. For SST-2 and QQP, teacher and conversion use a fixed part of the remaining training split. \Cref{app:protocol} gives every split and its size. The two models have 6 blocks each. DistilGPT2 \citep{sanh2019distilbert,radford2019language} is a decoder with causal attention whose classifier reads an end-of-text token appended to the tweet. BERT-base-uncased \citep{devlin2019bert} truncated to its first 6 blocks is an encoder with bidirectional attention whose classifier reads \cls. All are fine-tuned in FP16 mixed precision \citep{micikevicius2018mixed} and quantised after training to FP8 E4M3 \citep{micikevicius2022fp8}. The FP8 model is the teacher of every conversion, with a test gold accuracy on TweetEval-emotion of 0.747 for DistilGPT2 and 0.800 for the 6-block BERT (\cref{tab:main} for the GLUE tasks). Its arithmetic is already a finite circuit (\cref{app:fp8}), and the predicate circuit makes it small by keeping only the distinctions that the task uses.

\paragraph{Protocol.} Every conversion uses only the training split, restricted to the examples that the teacher classifies correctly (\cref{sec:fitset}). Choices made during the conversion use the validation split, separately for every task, network and form of selection, and the test split serves only for reporting. The main metric is test gold accuracy, and the second is agreement with the teacher on test, both as mean $\pm$ s.d.\ over 5 seeds of the conversion. Agreement on tweets of other TweetEval tasks, used in no conversion, is in \cref{app:ood}, and \cref{app:protocol} lists the settings.

\paragraph{Reference.} A bag of word pieces, a logistic regression over the counts of the teacher's word pieces, is fitted on the same training tweets with the teacher's labels. Its word weights are additive and cannot express logical combinations of words, which makes it the natural reference for an interpretable model of the same data. No earlier method turns a trained attention network into a hard-logic circuit, so we also compare with the closest earlier work, BiBERT \citep{qin2022bibert}, which binarises the weights, embeddings and activations of BERT to 1 bit. It is trained by its official code from the same 6-block teacher on the same splits, five seeds (\cref{app:bibert}).

\subsection{Results on five tasks}
\label{sec:exp-full}

\begin{table*}[t]
  \caption{Complete circuits on five tasks, fitted on the training examples that the teacher classifies correctly. Test gold accuracy and agreement with the teacher on test (GLUE: the GLUE validation split). Circuits and BiBERT: mean, with the s.d.\ over 5 seeds in small type. The bag of words is fitted on the same examples with the teacher's labels. BiBERT: \cref{app:bibert}. Splits: \cref{app:protocol}.}
  \label{tab:main}
  \centering
  \footnotesize
  \setlength{\tabcolsep}{1.9pt}
  \renewcommand{\sd}[1]{{\scriptsize$\pm$#1}}
  \begin{tabular}{@{}lcccccccccc@{}}
    \toprule
    & \multicolumn{2}{c}{TweetEval-emotion} & \multicolumn{2}{c}{SST-2} & \multicolumn{2}{c}{QQP} & \multicolumn{2}{c}{MRPC} & \multicolumn{2}{c}{CoLA} \\
    \cmidrule(lr){2-3}\cmidrule(lr){4-5}\cmidrule(lr){6-7}\cmidrule(lr){8-9}\cmidrule(l){10-11}
    Model & Gold & Agr. & Gold & Agr. & Gold & Agr. & Gold & Agr. & Gold & Agr. \\
    \midrule
    \multicolumn{11}{@{}l}{\emph{DistilGPT2 (decoder, 6 blocks)}} \\
    \quad teacher (FP8) & 0.747 & 1 & 0.883 & 1 & 0.801 & 1 & 0.765 & 1 & 0.742 & 1 \\
    \quad \orderSel & 0.677\sd{.008} & 0.710\sd{.005} & 0.777\sd{.011} & 0.802\sd{.010} & 0.707\sd{.003} & 0.742\sd{.005} & 0.671\sd{.024} & 0.689\sd{.014} & 0.670\sd{.013} & 0.739\sd{.009} \\
    \quad \ruleSel & 0.564\sd{.014} & 0.582\sd{.015} & 0.699\sd{.023} & 0.712\sd{.025} & 0.659\sd{.013} & 0.677\sd{.008} & 0.656\sd{.017} & 0.670\sd{.023} & 0.687\sd{.012} & 0.728\sd{.011} \\
    \quad bag of words & 0.640 & 0.653 & 0.763 & 0.788 & 0.670 & 0.697 & 0.681 & 0.667 & 0.629 & 0.720 \\
    \midrule
    \multicolumn{11}{@{}l}{\emph{BERT-base (encoder, first 6 blocks)}} \\
    \quad teacher (FP8) & 0.800 & 1 & 0.864 & 1 & 0.816 & 1 & 0.809 & 1 & 0.757 & 1 \\
    \quad \orderSel & 0.711\sd{.011} & 0.752\sd{.014} & 0.788\sd{.010} & 0.827\sd{.004} & 0.705\sd{.004} & 0.722\sd{.006} & 0.672\sd{.019} & 0.753\sd{.030} & 0.671\sd{.004} & 0.750\sd{.008} \\
    \quad \ruleSel & 0.618\sd{.050} & 0.633\sd{.056} & 0.747\sd{.016} & 0.769\sd{.018} & 0.656\sd{.010} & 0.655\sd{.012} & 0.680\sd{.015} & 0.772\sd{.016} & 0.686\sd{.005} & 0.754\sd{.006} \\
    \quad bag of words & 0.638 & 0.687 & 0.774 & 0.791 & 0.675 & 0.712 & 0.647 & 0.716 & 0.640 & 0.737 \\
    \midrule
    \multicolumn{11}{@{}l}{\emph{Closest earlier work: arithmetic network with 1-bit weights, embeddings and activations, from the 6-block BERT teacher}} \\
    \quad BiBERT & 0.739\sd{.010} & 0.803\sd{.013} & 0.798\sd{.011} & 0.837\sd{.005} & 0.725\sd{.005} & 0.752\sd{.008} & 0.694\sd{.008} & 0.802\sd{.006} & 0.662\sd{.027} & 0.774\sd{.031} \\
    \bottomrule
  \end{tabular}
\end{table*}

\Cref{tab:main} reports the complete circuits, in which no floating-point weight is left. On TweetEval-emotion and DistilGPT2, \orderSel\ reaches a test gold accuracy of 0.677 and an agreement of 0.710 over five seeds, above the bag of words fitted on the same tweets (0.640 and 0.653), at 47.9\,Mbit, 13.7 times smaller than the FP8 teacher (655.3\,Mbit, seed 0). With \ruleSel, where every step before the readout is a rule, it reaches 0.564 and 0.582 at 67.7\,Mbit. On the 6-block BERT, \orderSel\ reaches 0.711 and 0.752, 7.3 points of gold accuracy above the bag of words on the same tweets (0.638 and 0.687), at 29.3\,Mbit, 18.3 times smaller than the teacher (535.7\,Mbit), and \ruleSel\ reaches 0.618 and 0.633 at 49.0\,Mbit. Most of the size is the input table (\cref{tab:size}). On the four GLUE tasks, two with single sentences (SST-2, CoLA) and two with sentence pairs (QQP, MRPC), \orderSel\ reaches a mean test gold accuracy of 0.706 on DistilGPT2 and 0.709 on the 6-block BERT, 2.0 and 2.5 points above the bag of words fitted on the same examples (0.686 and 0.684), and \ruleSel\ reaches 0.675 and 0.692.

\paragraph{Closest earlier work.} BiBERT is a stricter reference than the bag of words, because it keeps arithmetic. It holds 66.5 million binary weights (75.0\,Mbit with its scales and remaining parameters in FP16), its linear layers take 42.5 million binary multiply--accumulates per token with real-valued scales and LayerNorm, and it is retrained end to end. Over the five tasks it reaches a test gold accuracy of 0.724 and an agreement of 0.794. The 6-block \orderSel\ circuit reaches 0.709 and 0.761 with table lookups and logic alone, at 29.2--29.3\,Mbit, 2.6 times smaller, without multipliers, scales or normalisation.

\begin{table*}[t]
  \caption{Cost of one classification, mean over the five tasks (circuits: \orderSel, also over five seeds). Latency: from the tokens to the class. M: measured. E: estimated from synthesis at 450--250\,MHz and typical to maximum board power, BiBERT as an idealised lower bound on the same devices. Size (whole / active vocabulary) and bits read per example by the same rule for every system. Throughput mode: board power / throughput. No circuit uses a DSP block or a multiplier. Prices: RTX PRO 6000 \$11.0k--17.9k and Alveo U55C \$4,747, each plus a host, Kria KV260 \$274. Details in \cref{app:hw}.}
  \label{tab:hw-main}
  \centering
  \footnotesize
  \setlength{\tabcolsep}{2.25pt}
  \begin{tabular}{@{}llcrrrrlc@{}}
    \toprule
    & & Size, Mbit & Read per & Latency, & \multicolumn{2}{c}{Energy per example} & & \\
    \cmidrule(lr){6-7}
    System & Device & whole / active & example & batch 1 & batch 1 & throughput & Arithmetic & Gold \\
    \midrule
    \multicolumn{9}{@{}l}{\emph{BERT-base, first 6 blocks}} \\
    teacher (M) & RTX PRO 6000 & 535.7 / 418.3 & 345.3\,Mbit & 460\,$\mu$s & 63.0\,mJ & 6.79\,mJ & FP16 & 0.809 \\
    BiBERT (E, bound) & Alveo U55C & 75.0 / 60.0 & 45.2\,Mbit & $\ge$\,1.82--3.28\,$\mu$s & $\ge$\,210--492\,$\mu$J & $\ge$\,210--492\,$\mu$J & 1-bit and real & 0.724 \\
    BiBERT (E, bound) & Kria KV260 & 75.0 / 60.0 & 45.2\,Mbit & $\ge$\,122.9\,$\mu$s & $\ge$\,922--1,843\,$\mu$J & $\ge$\,163--586\,$\mu$J & 1-bit and real & 0.724 \\
    circuit (E) & Alveo U55C & 29.2 / 11.2 & 0.43\,kbit & 1.10--1.98\,$\mu$s & 127--297\,$\mu$J & 19.3--45.2\,$\mu$J & none & 0.709 \\
    circuit (E) & Kria KV260 & 29.2 / 11.2 & 0.43\,kbit & 3.48--6.38\,$\mu$s & 26.1--95.7\,$\mu$J & 4.6--16.7\,$\mu$J & none & 0.709 \\
    \midrule
    \multicolumn{9}{@{}l}{\emph{DistilGPT2}} \\
    teacher (M) & RTX PRO 6000 & 655.3 / 439.3 & 340.6\,Mbit & 346\,$\mu$s & 50.3\,mJ & 8.16\,mJ & FP16 & 0.788 \\
    circuit (E) & Alveo U55C & 47.8 / 14.6 & 0.42\,kbit & 1.08--1.95\,$\mu$s & 125--292\,$\mu$J & 19.1--44.5\,$\mu$J & none & 0.700 \\
    circuit (E) & Kria KV260 & 47.8 / 14.6 & 0.42\,kbit & 3.33--6.12\,$\mu$s & 25.0--91.8\,$\mu$J & 4.4--16.0\,$\mu$J & none & 0.700 \\
    \bottomrule
  \end{tabular}
\end{table*}

\subsection{Hardware: bit-exact export and cost}
\label{sec:exp-hw}

\paragraph{Architecture and export.} The circuit is exported as a hardware description with one fixed engine per block, and the engines form a pipeline across tweets. Code maps, score tables or selection rules, counters and transition rules are lookup-table logic, the readout adds the integer votes of the predicates as they turn true, and the only memory is the input table, so no weight moves between memory and logic. Before export, the circuit is minimised exactly as a state machine \citep{huffman1954synthesis}, with the same states and decisions on every token (\cref{app:merge}). For all 100 circuits of \cref{tab:main} (five tasks, two networks, two forms of selection, five seeds), the synthesised lookup-table netlists, simulated on real inputs, agree with the exported description on every checked row, which agrees with the software circuit up to exact ties of the vote sums (\cref{app:hw}).

\paragraph{Measured and estimated.} The circuit side is estimated from logic synthesis to 6-input lookup tables with Yosys and ABC \citep{wolf2016yosys,brayton2010abc}, a clock bracketed at 250--450\,MHz by published results on the same device family and power from published per-resource energy coefficients, and the GPU side is measured with CUDA graphs, whose predictions equal the eager ones. The bound for BiBERT grants it more than these devices hold: one binary multiply--accumulate per lookup table and cycle, free real-valued operations, and the full memory bandwidth for weights that do not fit on chip (\cref{app:hw}).

\paragraph{Latency, energy and components.} Fixed logic answers fast and spends little energy (\cref{tab:hw-main}). Over the five tasks, the 6-block BERT circuit classifies an example in 1.10--1.98\,$\mu$s on an Alveo U55C card, 230--420 times faster than its teacher on an RTX PRO 6000 with CUDA graphs and, at the same clock, 1.7 times faster than the bound of BiBERT on the same card, with 19.3--45.2\,$\mu$J per example in throughput mode against at least 210\,$\mu$J. On a palm-sized Kria KV260 board of 7.5--15\,W, it takes 3.5--6.4\,$\mu$s and 26--96\,$\mu$J per example in a single stream, 660--2,400 times less energy than its teacher at batch 1. There the binary weights of BiBERT (66.5\,Mbit) exceed the 24.2\,Mbit of on-chip memory, so that 19\,Mbit of block weights stream from DRAM for every example (at least 122.9\,$\mu$s), whereas the circuit reads 0.43\,kbit of its input table and keeps all of its logic on chip. The DistilGPT2 circuit is as fast and as frugal, and the \ruleSel\ circuits are faster still (\cref{app:hw}).

\subsection{Conversion can stop at any block}
\label{sec:exp-stages}

\begin{table}[t]
  \caption{Conversion can stop at any block. Test gold accuracy and agreement with the teacher after the top $k$ blocks have become circuits, with the teacher's own blocks below (\orderSel, mean over five tasks and five seeds). $k=0$ is the FP8 teacher and $k=6$ the complete circuit of \cref{tab:main}. Per task: \cref{app:size}.}
  \label{tab:stages}
  \centering
  \footnotesize
  \setlength{\tabcolsep}{2.6pt}
  \begin{tabular}{@{}llccccccc@{}}
    \toprule
    & & \multicolumn{7}{c}{blocks converted, $k$} \\
    \cmidrule(l){3-9}
    Model & & 0 & 1 & 2 & 3 & 4 & 5 & 6 \\
    \midrule
    \multirow{2}{*}{DistilGPT2} & gold & .788 & .789 & .783 & .767 & .744 & .714 & .700 \\
     & agr. & 1 & .926 & .902 & .862 & .827 & .771 & .737 \\
    \midrule
    \multirow{2}{*}{BERT, 6 blocks} & gold & .809 & .785 & .770 & .750 & .721 & .716 & .709 \\
     & agr. & 1 & .916 & .878 & .837 & .788 & .776 & .761 \\
    \bottomrule
  \end{tabular}
\end{table}

Because the conversion runs from the output down on the teacher's own states, every stage of it is a complete classifier (\cref{tab:stages}). With its top block converted, DistilGPT2 keeps the gold accuracy of its teacher (0.789 against 0.788) at an agreement of 92.6\%, and the 6-block BERT agrees on 91.6\%. A deployment can thus stop at the block where accuracy and cost meet its needs, and the hardware figures of \cref{sec:exp-hw} are those of the complete circuit.

\subsection{Reading the circuit}
\label{sec:exp-read}

Take the test tweet ``So exhausted I could cry [loudly crying face emoji]'', whose gold label is sadness. The byte-level tokenizer of DistilGPT2 splits the emoji into two tokens. The \orderSel\ circuit of DistilGPT2 (seed 0) answers sadness, 495 vote units ahead of joy. In the last block $\Gamma_5$, head 0 of the readout token selects every token of this short tweet, and of the 61 predicates that hold, the three that speak most for sadness against joy are predicates on these keys: head 0 selects a key with lexical code $c_x$ (votes $+75$ for sadness, $-19$ for joy), a key with lexical code $c_y$ ($+37$ / $-50$) and a key in state $s_a$ ($+79$ / $-3$).
The key with code $c_x$ is ``exhausted'', which shares this code with ``feel'', ``feeling'' and ``alone'', and the key with code $c_y$ is ``cry'', which shares it with ``bad'' and ``lack''. At that interface ``exhausted'' is also in state $s_a$, which the circuit gives to ``heart'', ``miss'' and ``lost'' as well. Every predicate of the decision is thus named by the words that share its code or state, and the class follows from a short list of facts about the keys that attention selects.

\subsection{Imitating the teacher where it is right}
\label{sec:exp-ablation}

Against a conversion on all training tweets, which also copies the teacher's errors, fitting only where the teacher is right raises the test gold accuracy of DistilGPT2 at seed 0 from 0.664 to 0.683 with \orderSel\ and from 0.524 to 0.548 with \ruleSel\ (\cref{tab:ablation} in \cref{app:fitset}). The 6-block BERT teacher errs on only 0.3\% of its training tweets.

\section{Discussion and Future Work}
\label{sec:discussion}

This study converts every block and every operation of a pretrained decoder and a pretrained encoder, 6 blocks each, on five text-classification tasks with single sentences and sentence pairs, and verifies the exported hardware bit for bit. Future work will apply the same procedure to other networks built from maps between token features, products of two activations, competitions and sums over positions, among them decoders that generate text and larger encoders, and will place a tokenizer built in logic beside the circuit, so that its microsecond latency carries from raw text to class on a single small board.

\section{Conclusion}
\label{sec:conclusion}

We turned a trained attention network as a whole into a hard-logic predicate circuit in which every map is a minimised truth table, converted block by block from the output down, so that the errors of its blocks add rather than compound, and fitted only where the teacher is right.

On five text-classification tasks and two architectures, the circuit is above a bag of words fitted on the same examples on average and comes within 1.5 points of BiBERT, the closest earlier work, with logic alone and 2.6 times fewer bits. Its predicates and rules can be read decision by decision, and the conversion can stop at any block.

By a synthesis-based estimate, the bit-exact hardware classifies an example in 1.1--2.0\,$\mu$s on an FPGA card and in 3.3--6.4\,$\mu$s on a palm-sized board, with 550--2,400 times less energy than the network on a GPU.

\section*{Software and Data}
The experiments use the public TweetEval and GLUE benchmarks and the public DistilGPT2 and BERT-base checkpoints.

\section*{Impact Statement}
This paper presents work whose goal is to advance the field of Machine Learning. Converting networks into readable logic circuits can make their decisions easier to audit and cheaper to run. There are many potential societal consequences of our work, none of which we feel must be specifically highlighted here.

\section*{AI Use Statement}
An AI assistant (Claude, Anthropic) assisted with engineering code, data organisation and the writing of the paper. The research ideas and the method are the authors' own; the authors checked all results and the text and take full responsibility for them.

\bibliography{refs}
\bibliographystyle{icml2027}

\newpage
\appendix
\crefalias{section}{appendix}
\onecolumn

\section{Reproducibility: data, models and metrics}
\label{app:protocol}

\begin{table}[H]
  \caption{Data, teachers, fit sets and metrics.}
  \label{tab:protocol-full}
  \centering
  \small
  \begin{tabular}{@{}p{0.24\textwidth}p{0.7\textwidth}@{}}
    \toprule
    Item & Setting \\
    \midrule
    Main task & TweetEval-emotion, 4 classes, official splits of 3,257 training, 374 validation and 1,421 test tweets \\
    GLUE tasks & SST-2, QQP, MRPC and CoLA, 2 classes each. Test: the GLUE validation split (872 examples, a fixed 5,000 of the QQP pairs, 408, 1,043). Validation: a stratified part of the GLUE training split (2,000, 2,000, 400, 1,000). Training: the rest of the GLUE training split, of which SST-2 and QQP use the first 15,000 rows (MRPC 3,268, CoLA 7,551). A sentence pair is one sequence with the separator token of the network, maximum length 64 \\
    GLUE teachers & per task, the protocol of the same network on TweetEval-emotion below, with the learning rate chosen on validation macro-F1 among $2\cdot10^{-5}$, $5\cdot10^{-5}$ and $10^{-4}$ (6-block BERT) or $5\cdot10^{-5}$, $10^{-4}$ and $2\cdot10^{-4}$ (DistilGPT2). Test gold accuracies in \cref{tab:main} \\
    Teacher training & FP16 mixed precision with FP32 master weights and a dynamic loss scale, batch 128, weight decay 0.01, warm-up 10\%, class-weighted cross-entropy, maximum length 64, seed 42, epoch selected on validation \\
    DistilGPT2 teacher & DistilGPT2, 6 blocks of 12 heads, causal attention, an end-of-text token appended as the readout token and a new padding token. Learning rate $2\cdot10^{-4}$ and 10 epochs, the learning rate chosen on validation macro-F1 among $5\cdot10^{-5}$, $10^{-4}$ and $2\cdot10^{-4}$ \\
    6-block BERT teacher & the first 6 blocks of BERT-base-uncased with their pretrained weights and a new classifier. Learning rate $5\cdot10^{-5}$ and 10 epochs, the learning rate chosen on validation macro-F1 \\
    FP8 teacher & post-training E4M3 quantisation of every linear layer and of both attention products, weights scaled per output channel, operands of every product scaled dynamically per tensor (per head in attention), FP32 accumulation, softmax, LayerNorm and GELU in FP32 \\
    Teacher accuracy & gold accuracy on training / test: 0.931 / 0.747 (DistilGPT2) and 0.997 / 0.800 (6-block BERT) \\
    Fit set & training tweets that the FP8 teacher classifies correctly: 3,032 (DistilGPT2) and 3,247 (6-block BERT) of 3,257. Ablation: all 3,257 training tweets with the teacher's labels. GLUE, DistilGPT2 / 6-block BERT: SST-2 14,931 / 14,924 and QQP 14,470 / 14,887 of 15,000, MRPC 2,909 / 3,254 of 3,268, CoLA 7,097 / 7,485 of 7,551 \\
    Selection and reporting & choices during the conversion on the validation split only (on its teacher-correct tweets when the fit set is teacher-correct), the test split for reporting only \\
    Out-of-domain tweets & 2,000 tweets of the other TweetEval tasks, deduplicated against the emotion task, used in no conversion, agreement only (\cref{app:ood}) \\
    Metrics & test gold accuracy, agreement with the FP8 teacher on test, mean $\pm$ s.d.\ over 5 seeds of the conversion (fit-set ablation: seed 0) \\
    Bag of words & counts of the teacher's word pieces without readout and separator tokens, standardised per column, multinomial logistic regression on the teacher's labels with an L2 penalty chosen by validation cross-entropy, fitted on the same tweets as the circuit \\
    \bottomrule
  \end{tabular}
\end{table}

\paragraph{BiBERT.}\label{app:bibert} The closest earlier work, BiBERT \citep{qin2022bibert}, runs by its official code (commit \texttt{014399cb}, unchanged since its release) in its fully binarised setting: 1-bit weights and word embeddings, $\mathrm{sign}(w-\bar w)$ scaled by the mean absolute value of each output row, 1-bit activations, and its bi-attention, which replaces the softmax by a binary attention map. Position and token-type embeddings, biases, LayerNorm and the classifier stay real-valued. The student starts from the fine-tuned 6-block BERT teacher of each task, the same model whose FP8 version is the teacher of the circuits, and learns from it by BiBERT's loss: soft cross-entropy on the teacher's logits, half the cross-entropy on the gold labels, direction-matching distillation on the query, key and value similarity matrices, and the distance between the normalised hidden states of the embedding layer and of every block. It is trained on the whole training split of each task (for SST-2 and QQP, the same 15,000 rows), without data augmentation, with the code's optimiser (BertAdam, linear warm-up over 10\% of the steps, weight decay 0.01) and batch 16. Learning rate and epochs follow the official scripts and BiBERT's appendix: $10^{-4}$ for SST-2 (10 epochs), QQP (5) and CoLA (50), and $2\cdot10^{-4}$ for MRPC (20); TweetEval-emotion, which is not part of GLUE, takes the MRPC row, the GLUE task closest in size (20 epochs), and runs through the single-sentence path of the code with its four labels. Inputs are cut at 64 tokens like the teachers' (the official code pads MRPC and QQP to 128). The checkpoint is chosen by the code's own rule, the best validation accuracy (Matthews correlation for CoLA) over evaluations every 200 steps (500 for CoLA, 1,000 for QQP), on the same validation split as the circuits, and reported once on test as test gold accuracy and agreement with the FP8 teacher, seeds 0--4. Its size counts the 66,498,048 binary weights of its 38 binary matrices, word embeddings included, at 1 bit, and their 72,762 scales and the remaining 0.46 million real-valued parameters at 16 bits, 75.0\,Mbit in all.

\section{Reproducibility: implementation details}
\label{app:impl}

This appendix gives every step and setting of the conversion as the experiments run it, for both networks, every task and both forms of selection. \Cref{tab:impl} collects the numbers and \cref{tab:thresholds} the thresholds chosen for each group.

\paragraph{Conventions.} Sequences have at most 64 tokens, and the readout token is at position 0. The circuit stores a DistilGPT2 sequence reversed, so that its final token is at position 0 and every token selects among the keys at or after it in the stored order. The teacher runs in FP32 with TF32 disabled, without dropout and without segment embeddings, as in its fine-tuning. Every interface has 16 states for the readout token and 64 for the other tokens, 80 in all. The lexical codes number 257, 256 for the word pieces and one for the readout token.

\paragraph{Fit set, validation and seeds.} The fit set holds the training rows that the FP8 teacher classifies correctly (\cref{sec:fitset}), and every target of the conversion is the teacher's class or a state derived from it. A fifth of the fit tweets, drawn with the seed, is held out inside the fit set. It chooses the penalty of the transition surrogates, forms the second half of the co-firing test of predicates, and accepts or rejects the exception rules. The validation split chooses the penalty of the vote readout and the thresholds $\pi$ and $s_{\min}$, and the test split serves reporting only. The seed fixes the held-out fifth, every $k$-means initialisation, the seed rows of the rule learner and the query--key pairs sampled for \ruleSel. It changes neither the fit set nor the teacher.

\paragraph{Lexical codes.} The codes are computed once, from the output of the embedding. Two ridge regressions on the standardised output of a word piece predict its centred log attention weight from the readout token in every head of every block, and the class of its sequence. Each prediction is scaled to unit total variance, and $k$-means clusters the concatenation into 256 codes.

\paragraph{Interfaces.} At the top interface, a softmax probe on the readout token predicts the teacher's class, and $k$-means clusters its class probabilities into 16 states. All other tokens share one state there. Interface $m$ below is fitted on the teacher, with the states that interface $m+1$ assigns to the output of block $m$ as targets. A readout token takes the nearest of 16 centroids of a probe's probabilities of its target state. Another token takes the nearest of 64 centroids of three concatenated parts: a probe's probabilities of its own target state; a probe of the target state of the readout token, fitted with each token weighted by the mean attention of the readout token to it in block $m$; and a ridge prediction of its centred log attention weight from the readout token in each head of block $m$, scaled to the total variance of the two other parts.

\paragraph{Predicate bank.} The literals of a token are its own state, $[s_i=x]$ for the 80 states, and for every head and every state or lexical code $x$ whether at least one and whether at least two of the selected keys carry $x$. These are read from one saturating 2-bit counter per head and per state or code, 4,044 for 12 heads. ORs across heads, whether some head selects at least one or at least two keys carrying $x$, are added. Literals that hold on fewer than 10 rows are dropped. Literals whose co-firing Jaccard index is at least 0.95 both on the fit rows and on the held-out rows are merged into the most frequent of them, greedily in order of support. Weak literals, which hold on 3 to 9 rows, are grouped into ORs. Around a seed literal, candidates whose distribution of target states lies within a Jensen--Shannon distance of 0.2 join while a chi-square homogeneity test against the running group gives $p>0.01$, with at most 32 members out of at most 128 candidates, and an OR is kept when it holds on at least 10 rows.

\paragraph{Rule learner.} One learner serves the transitions of the readout token, the exceptions of the other tokens and \ruleSel. A cube grows from the empty cube by the literal of a seed row that has the best FOIL gain among those that remove at least one covered row of another class and keep at least $s_{\min}$ rows of the seed's class. It stops at precision $\pi$, when no row of another class is left, or at 6 literals. It is then pruned by dropping literals while its precision stays at least $\pi$, and kept when it covers at least $s_{\min}$ rows with precision at least $\pi$. Each class draws 256 seed rows per round, in 4 rounds, from its uncovered rows. A greedy cover then adds cubes while the score $(n_{\text{new correct}}-n_{\text{new wrong}})/(1+0.05\,\ell)$, with $\ell$ literals, is positive. A row on which the cubes of one class fire takes that class. Where several classes fire, a conflict table for that pattern of firing classes decides, or else the cube with the highest Wilson lower bound ($z=1.96$). Where none fires, a local default for the row's own state decides, or else the majority class. A conflict entry or a local default needs at least $s_{\min}$ rows, and a conflict entry is kept only where it beats the prediction it replaces. Where exceptions are used, one round of cubes seeded by wrongly predicted rows is kept when it corrects at least $\max(2,\lfloor s_{\min}/2\rfloor)$ fit rows and gains on the held-out rows, and exceptions override every other rule.

\paragraph{Attention.} For each head a dense fit comes first. The score of a pair is $T_h(s_i,s_j)+V_h(s_i,c_j)+R_h(j-i)+U_h(s_i)\,[j\text{ is the last token}]$, with 127 relative positions, initialised at zero and fitted by Adam (learning rate 0.1, 200 full-batch steps) to the cross-entropy between the current network's attention probabilities in block $m$ and the softmax of the scores. In DistilGPT2, keys outside the causal range get $-10^4$. The query rows of $T_h$ and $V_h$, centred and weighted by frequency, are clustered into 16 query codes $Q_h$, the key columns of $T_h$ into 16 state codes $K^1_h$ and those of $V_h$ into 16 lexical codes $K^2_h$, and 21 rounds of alternating co-clustering refine them. $U_h$ becomes its frequency-weighted mean over each query code. With the codes fixed, the tables $S^1_h$ ($16\times16$), $S^2_h$ ($16\times16$), $U_h$ (16 entries) and $R_h$ (127) are refitted in the same way and rounded to integers after multiplication by 2. \OrderSel\ keeps every key whose integer score is at least the row maximum minus $\delta=3$, the margin of 1.5 times the scale of 2, ties included, with no fixed number of keys.

For \ruleSel, every query--key pair is a row over 69 one-hot columns: the query code, the two key codes, the relative position in 17 bins ($\le-8$, $-7,\dots,7$, $\ge8$) and a four-valued flag that marks the readout token and the last token. The target is the network's own hard selection, a key within 1.5 nats of the row maximum of the log attention. Pairs are sampled with probability $3\cdot300{,}000/n_{\text{pairs}}$ and capped at 300,000 per head, and DistilGPT2 keeps only the causal pairs. The rule learner runs with two classes and the group's $\pi$ and $s_{\min}$ on the queries outside the held-out fifth, with exceptions checked on the held-out queries and without local defaults. The rules of each head are compiled into a one-bit table over $16\times16\times16\times17\times4$ inputs, and the integer score tables are then not part of the circuit.

\paragraph{Transitions and state merging.} The readout token's next state comes from the rule learner on its predicate bank, with local defaults for its own state and one round of exceptions. Another token takes the majority next state of its own state, unless exception cubes learned on a sample of 40,000 rows override it. Their positive rows are the rows of a target that the default misses, a conflict table resolves overlaps between them, and there is no further round of exceptions. A target that never varies becomes a constant. After the last block, the rules and votes of every circuit block are refitted twice on the states that the complete circuit chain itself produces, with the states each block was built to produce as targets. The rows of these versions are pooled, so the support thresholds count pooled rows, and the attention tables are not refitted. From the second stage on, two states of the same kind whose query and state-key codes in the block above differ in at most 4 of the $2H$ codes are candidates for merging. They are tested in order of that difference and then of rarity, at most 150 pairs, and a pair merges into the more frequent state when the swap, with the attention recomputed, changes the output of the block above on fewer than 2\% of the rows of the tweets that contain it.

\paragraph{Readout.} The vote readout reads the bank of the readout token in the last block: its own state and the at-least-one and at-least-two counts of every head, without the ORs across heads, keeping the literals that hold on at least 5 fit rows. A multinomial logistic regression, fitted by L-BFGS (300 iterations) with an unpenalised bias and an L2 penalty from $\{10^{-4},10^{-3},10^{-2}\}$ chosen by validation cross-entropy, gives the votes. They are quantised per matrix to 8-bit integers with the scale $\max|W|/127$, the biases to 16-bit integers, and the readout is refitted after every block and after the final refitting of the chain.

\paragraph{Thresholds.} The precision threshold $\pi$ and the support threshold $s_{\min}$ of the rule learner are chosen for each task, network and form of selection. Each of the six settings $\pi\in\{0.7,0.8,0.9\}\times s_{\min}\in\{1,4\}$ is converted with seed 0 on a tenth of the fit set, and the setting with the highest validation gold accuracy of the complete circuit is kept. A tie keeps the default $(\pi,s_{\min})=(0.8,1)$ when it is among the best, and otherwise the setting with fewer rule literals wins. The main runs use $\pi$ and $10\,s_{\min}$, for a fit set ten times as large (\cref{tab:thresholds}).

\paragraph{Size accounting.} Every integer table, the code maps $Q_h$ (80 entries), $K^1_h$ (80) and $K^2_h$ (257) and the score tables $S^1_h$, $S^2_h$, $U_h$ and $R_h$, is stored with as many bits per entry as its range needs. Each of its output bits is minimised exactly, by the prime implicants of Quine--McCluskey and a minimum cover by integer programming, with the fewest cubes and then the fewest literals and the unused codes as don't-cares. For \ruleSel\ the compiled selection tables, $278{,}528\,H$ bits per block, replace the score tables. The votes count 8 bits per weight, 16 per bias and 32 for the scale, and the 16-entry class table of the top interface is counted as well. The input table $\Lambda$ maps each token and each of 63 positions to a 6-bit state and a 9-bit lexical code, and the totals in this paper count it over the whole vocabulary. The transition rules, their conflict tables and local defaults are logic, counted in literals and not in bits (\cref{tab:size}).

\begin{table}[H]
  \caption{Settings of the conversion, identical for both networks, every task and both forms of selection unless stated.}
  \label{tab:impl}
  \centering
  \small
  \begin{tabular}{@{}p{0.2\textwidth}p{0.74\textwidth}@{}}
    \toprule
    Part & Setting \\
    \midrule
    Symbols & 16 states of the readout token and 64 of the other tokens per interface; 256 lexical codes and one for the readout token; at most 64 tokens \\
    Probes and clustering & softmax probes on standardised features, L-BFGS, L2 penalty $10^{-3}$, 100 iterations; ridge penalty $10^{-2}$; $k$-means with 30 Lloyd iterations \\
    Predicates & support $\ge10$ rows; merge at co-firing Jaccard $\ge0.95$ on fit and held-out rows; ORs of weak literals (3--9 rows) at Jensen--Shannon distance $\le0.2$ and chi-square $p>0.01$, at most 32 of 128 candidates \\
    Rule learner & precision $\pi$ and support $s_{\min}$ (\cref{tab:thresholds}); at most 6 literals per cube; 256 seed rows per class and round, 4 rounds; cover score $(n_{\text{correct}}-n_{\text{wrong}})/(1+0.05\,\ell)$; Wilson bound $z=1.96$ for ranking; one round of exceptions \\
    Attention & dense fit by Adam, learning rate 0.1, 200 steps; per head 16 query, 16 state-key and 16 lexical-key codes, 21 co-clustering rounds; 127 relative positions; integer scale 2; \orderSel\ margin $\delta=3$ \\
    \RuleSel & 69 one-hot columns, 17 relative-position bins; target within 1.5 nats of the row maximum; at most 300,000 sampled pairs per head \\
    Transitions & exception cubes of the other tokens on 40,000 sampled rows; two rounds of refitting on the complete chain's own states after the last block; state merging for at most 4 of $2H$ differing codes, at most 150 pairs, output change below 2\% \\
    Readout & literals on $\ge5$ fit rows; L2 penalty from $\{10^{-4},10^{-3},10^{-2}\}$ on validation, L-BFGS, 300 iterations; 8-bit votes, 16-bit biases \\
    Threshold search & $\pi\in\{0.7,0.8,0.9\}$, $s_{\min}\in\{1,4\}$, seed 0, a tenth of the fit set, validation gold accuracy; main runs at $10\,s_{\min}$ \\
    \bottomrule
  \end{tabular}
\end{table}

\begin{table}[H]
  \caption{Thresholds $(\pi, s_{\min})$ of the main runs, chosen on validation (\cref{app:impl}).}
  \label{tab:thresholds}
  \centering
  \small
  \begin{tabular}{@{}lcccc@{}}
    \toprule
    & \multicolumn{2}{c}{DistilGPT2} & \multicolumn{2}{c}{BERT-base, 6 blocks} \\
    \cmidrule(lr){2-3}\cmidrule(l){4-5}
    Task & \orderSel & \ruleSel & \orderSel & \ruleSel \\
    \midrule
    TweetEval-emotion & (0.8, 40) & (0.8, 10) & (0.8, 10) & (0.7, 10) \\
    SST-2 & (0.7, 40) & (0.8, 40) & (0.7, 40) & (0.7, 40) \\
    QQP & (0.8, 40) & (0.7, 40) & (0.7, 40) & (0.7, 40) \\
    MRPC & (0.8, 10) & (0.7, 40) & (0.8, 40) & (0.7, 40) \\
    CoLA & (0.8, 10) & (0.7, 40) & (0.7, 40) & (0.7, 40) \\
    \bottomrule
  \end{tabular}
\end{table}

\section{Fitting only where the teacher is right}
\label{app:fitset}

Teacher errors on the training split. The DistilGPT2 teacher of \cref{sec:experiments} is right on 93.1\% of its training tweets and on 74.7\% of the test tweets, and the 6-block BERT teacher on 99.7\% and 80.0\%.

\begin{table}[H]
  \caption{Fitting only where the teacher is right. The same conversion fitted on the training tweets that the teacher classifies correctly (sizes in \cref{app:protocol}) or on all training tweets with the teacher's labels. Test gold accuracy and agreement with the teacher on test, seed 0.}
  \label{tab:ablation}
  \centering
  \small
  \setlength{\tabcolsep}{3pt}
  \begin{tabular}{@{}llcccc@{}}
    \toprule
    & & \multicolumn{2}{c}{DistilGPT2} & \multicolumn{2}{c}{6-block BERT} \\
    \cmidrule(lr){3-4}\cmidrule(lr){5-6}
    & Fit set & Gold & Agr. & Gold & Agr. \\
    \midrule
    \multirow{2}{*}{\orderSel} & teacher-correct & 0.683 & 0.710 & 0.723 & 0.768 \\
     & all & 0.664 & 0.720 & 0.727 & 0.774 \\
    \midrule
    \multirow{2}{*}{\ruleSel} & teacher-correct & 0.548 & 0.564 & 0.543 & 0.549 \\
     & all & 0.524 & 0.569 & 0.596 & 0.616 \\
    \midrule
    \multirow{2}{*}{bag of words} & teacher-correct & 0.640 & 0.653 & 0.638 & 0.687 \\
     & all & 0.638 & 0.655 & 0.639 & 0.689 \\
    \bottomrule
  \end{tabular}
\end{table}

\Cref{tab:ablation} repeats the conversion on all training tweets, so that the circuit also copies the teacher's errors on them. On DistilGPT2 at seed 0, the teacher-correct fit set raises gold accuracy from 0.664 to 0.683 with \orderSel\ and from 0.524 to 0.548 with \ruleSel, while the agreement with the teacher on test is 0.710 and 0.564 against 0.720 and 0.569 on all training tweets. Fitting only where the teacher is right thus gains about two points of gold accuracy for less than one point of agreement. The bag of words changes by at most 0.002 in either measure. The 6-block BERT teacher errs on only 0.3\% of its training tweets, so the two fit sets nearly coincide. With these tweets in the fit set, its \orderSel\ circuit changes by at most 0.006 in either measure and its \ruleSel\ circuit by 0.054 in gold accuracy and 0.067 in agreement, about one standard deviation of \ruleSel\ over seeds (0.050 and 0.056, \cref{tab:main}).

\section{The FP8 teacher and its multiplier}
\label{app:fp8}

The product of two E4M3 codes splits into a sign, an exponent sum and a $4\times4$ significand product, each a truth table over at most 8 input bits. Every output bit of these complete tables is minimised exactly. The resulting multiplier has 240 AND terms with 1,222 literals and shows 0 mismatches against the exact product over all $254^2=64{,}516$ pairs of finite codes. Every attention product of the FP8 teachers is therefore available exactly from this circuit, and FP32 accumulation follows it as in the teachers. The FP8 DistilGPT2 teacher reaches a test gold accuracy of 0.747, against 0.746 in FP16, and the FP8 6-block BERT teacher 0.800, against 0.799 in FP16.

\section{Out-of-domain agreement}
\label{app:ood}

\begin{table}[H]
  \caption{Agreement with the teacher on the out-of-domain tweets, which enter no conversion. Main rows: mean $\pm$ s.d.\ over 5 seeds; ablation rows: seed 0.}
  \label{tab:ood}
  \centering
  \small
  \begin{tabular}{@{}llcc@{}}
    \toprule
    Fit set & Model & DistilGPT2 & BERT-base, 6 blocks \\
    \midrule
    \multirow{3}{*}{teacher-correct (main)} & \orderSel & 0.623\,$\pm$\,0.010 & 0.627\,$\pm$\,0.015 \\
     & \ruleSel & 0.476\,$\pm$\,0.046 & 0.484\,$\pm$\,0.040 \\
     & bag of words & 0.428 & 0.473 \\
    \midrule
    \multirow{3}{*}{all training tweets (ablation)} & \orderSel & 0.646 & 0.658 \\
     & \ruleSel & 0.472 & 0.451 \\
     & bag of words & 0.416 & 0.476 \\
    \bottomrule
  \end{tabular}
\end{table}

On these tweets, which come from other tasks, the circuits fitted on the teacher-correct training tweets agree with the teacher on 0.623 (\orderSel) and 0.476 (\ruleSel) for DistilGPT2 and on 0.627 and 0.484 for the 6-block BERT over five seeds, against 0.428 and 0.473 for the bags of words fitted on the same tweets. At seed 0, fitting on all training tweets instead gives 0.646 and 0.472 for DistilGPT2 (teacher-correct: 0.623 and 0.500) and 0.658 and 0.451 for the 6-block BERT (teacher-correct: 0.634 and 0.443).

\section{Hardware comparison: what is measured and what is estimated}
\label{app:hw}

\begin{table}[H]
  \caption{Latency and power for one tweet. Circuits: \orderSel, estimated (E) from synthesis and a stated performance and power model, latency of the logic as the mean over the test tweets at 250--450\,MHz, on the U55C the same for all five seeds and on the KV260 the range over the five seeds. Card and module power: typical / maximum, sourced (S) from the vendor. GPUs: the teacher at batch 1, measured (M), mean latency, in eager PyTorch the better of FP16 and FP32 with board power over both, with CUDA graphs FP16 (best tweet in parentheses), board power without the host. Gold accuracy (mean over five seeds): DistilGPT2 circuit 0.677, teacher 0.747; 6-block BERT circuit 0.711, teacher 0.800.}
  \label{tab:hw}
  \centering
  \small
  \setlength{\tabcolsep}{5pt}
  \begin{tabular}{@{}llrll@{}}
    \toprule
    Model & System & Latency, one tweet & Power & Form \\
    \midrule
    \multirow{5}{*}{DistilGPT2} & teacher on an RTX 3090, eager (M) & 2.60\,ms & board 136--181\,W (M) & card in a desktop \\
    & teacher on an RTX 5090, eager (M) & 2.21\,ms & board 119--160\,W (M) & card in a desktop \\
    & teacher on an RTX 5090, CUDA graphs (M) & 0.89\,ms (0.29) & board 161\,W (M) & card in a desktop \\
    & circuit on an Alveo U55C card (E) & 0.97--1.75\,$\mu$s & card 115 / 150\,W (S) & card in a host \\
    & circuit on a Kria KV260 board (E) & 1.9--3.7\,$\mu$s & module 7.5 / 15\,W (S) & palm-sized board \\
    \midrule
    \multirow{6}{*}{\shortstack[l]{BERT-base,\\6 blocks}} & teacher on an RTX 3090, eager (M) & 2.28\,ms & board 142--196\,W (M) & card in a desktop \\
    & teacher on an RTX 5090, eager (M) & 1.89\,ms & board 131--188\,W (M) & card in a desktop \\
    & teacher on an RTX PRO 6000, eager (M) & 1.47\,ms & board 118--154\,W (M) & card in a server \\
    & teacher on an RTX PRO 6000, CUDA graphs (M) & 0.46\,ms (0.43) & board 138\,W (M) & card in a server \\
    & circuit on an Alveo U55C card (E) & 1.08--1.94\,$\mu$s & card 115 / 150\,W (S) & card in a host \\
    & circuit on a Kria KV260 board (E) & 5.1--9.1\,$\mu$s & module 7.5 / 15\,W (S) & palm-sized board \\
    \bottomrule
  \end{tabular}
\end{table}

Within a block, the keys of a tweet stream past parallel query units. Before export, states that the circuit's own tables cannot produce are removed, and states that nothing above can tell apart are merged. On every token of the fit set and of the validation and test splits, the minimised circuit reaches the same states and decisions, its latency and throughput are unchanged, and its lookup tables change by less than 1\% (\cref{app:merge}). All circuit numbers are for the main-line circuits of DistilGPT2 and the 6-block BERT-base (\orderSel\ unless stated), after state-machine merging (\cref{app:merge}). On the U55C, the number of units, the latency, the throughput and the energy at board power are the same for all five seeds, while the area, the dynamic energy and the number of units that fit a KV260 vary slightly with the seed (\cref{tab:hw-seeds}). The hardware sums the integer votes exactly and takes the first maximum on a tie, whereas the reported accuracies come from the software circuit, which sums in floating point; the two differ only at an exact tie of the vote sums, which changes one test tweet in one of the 20 circuits (6-block BERT), and the integer value is the exact one for the circuit as specified.

\paragraph{Five tasks.} \Cref{tab:hw-tasks} gives the comparison of \cref{tab:hw-main} for every task, \cref{tab:hw-rule-tasks} the \ruleSel\ circuits, \cref{tab:hw-comp-tasks} the components of every circuit and \cref{tab:gpu4090-tasks} the teachers on an RTX 4090. Its GPU numbers were measured on an RTX PRO 6000 with the teacher's configuration and random weights. With the trained weights, the CUDA-graph latency agrees within 4\% and the board power is the same or up to 16\% higher, so the teacher energies are on the low side. With CUDA graphs, this GPU matches the measurement of \cref{tab:hw} on another host with the same GPU model (0.454 against 0.462\,ms for the 6-block BERT on TweetEval-emotion), while eager PyTorch at batch 1 is bound by the host processor that launches the kernels (3.35 against 1.47\,ms), which is why the main text reports CUDA graphs. On this host, eager PyTorch at batch 1 takes 3.31\,ms for the 6-block BERT and 4.22\,ms for DistilGPT2, mean over the five tasks, and an RTX 4090 with CUDA graphs takes 527 and 408\,$\mu$s. On the KV260, the latency of a circuit includes the lookup of its input table in board DRAM (0.1--0.3\,$\mu$s). Bits read per example count what is fetched from memory for one example, by the same rule for every system: every weight held in memory once (all weights of the teacher and of BiBERT), and one row per token of every table indexed by the input (the word-piece and position embeddings of the teacher and of BiBERT, the input table of a circuit), at the mean test length of the task. The other tables of a circuit are lookup-table logic and are not fetched. The latency of a circuit is the same for all five seeds, except for DistilGPT2 with \orderSel\ on MRPC and CoLA, where the largest design that fits one U55C has 32 units per block for some seeds and 63 for others. Over all 100 circuits, the exported hardware differs from the software circuit only at exact ties of the vote sums, so that its test gold accuracy equals that of \cref{tab:main} within one example for every \orderSel\ circuit and within five QQP pairs with \ruleSel. With \ruleSel, the circuits take 0.70--1.26\,$\mu$s (6-block BERT) and 0.69--1.24\,$\mu$s (DistilGPT2) per example on the U55C, mean over the five tasks.

\paragraph{The bound for BiBERT.} No 1-bit accelerator is built. The bound counts the operations of the inference graph of BiBERT's official code, its binary linear layers (query, key, value, attention output, both feed-forward layers and the pooler) and binary word embedding, 66.5\,Mbit of binary weights in all, and its binary attention map, 941 million binary multiply--accumulates per TweetEval-emotion test tweet on average. Every lookup table of the device performs one binary multiply--accumulate per cycle with free accumulation, the real-valued scales, LayerNorm, GELU and remaining embeddings cost nothing, and the binary weights stay on chip where they fit. On the U55C they fit, and compute sets the bound. On the KV260 the 43.1\,Mbit of the blocks and the pooler exceed its 24.2\,Mbit of on-chip memory, and the other 18.9\,Mbit stream from its DDR4 memory at the full 19.2\,GB/s for every example, which sets the bound at 122.9\,$\mu$s per example at batch 1. Clocks and board power are those of the circuits, and energy follows from them in the same way.

\begin{table}[H]
  \caption{Cost of one classification per task, as in \cref{tab:hw-main}: latency at batch 1, energy per example at batch 1 and in throughput mode, and test gold accuracy (circuits and BiBERT: mean over five seeds, \cref{tab:main}). M: measured, E: estimated. GPU: CUDA graphs at batch 1, and at batch 256 the better of eager PyTorch and CUDA graphs. Circuits: \orderSel, mean over five seeds.}
  \label{tab:hw-tasks}
  \centering
  \small
  \setlength{\tabcolsep}{4pt}
  \begin{tabular}{@{}llrrrc@{}}
    \toprule
    Task & System & Latency, batch 1 & Energy, batch 1 & Energy, throughput & Gold \\
    \midrule
    \multirow{8}{*}{TweetEval-emotion} & BERT teacher, RTX PRO 6000 (M) & 454\,$\mu$s & 60.7\,mJ & 6.07\,mJ & 0.800 \\
     & BiBERT, Alveo U55C (E, bound) & $\ge$\,1.60--2.89\,$\mu$s & $\ge$\,184--433\,$\mu$J & $\ge$\,184--433\,$\mu$J & 0.739 \\
     & BiBERT, Kria KV260 (E, bound) & $\ge$\,122.9\,$\mu$s & $\ge$\,922--1,843\,$\mu$J & $\ge$\,143--515\,$\mu$J & 0.739 \\
     & BERT circuit, Alveo U55C (E) & 1.08--1.94\,$\mu$s & 124--291\,$\mu$J & 19.3--45.2\,$\mu$J & 0.711 \\
     & BERT circuit, Kria KV260 (E) & 5.18--9.44\,$\mu$s & 38.8--141.6\,$\mu$J & 7.2--26.0\,$\mu$J & 0.711 \\
     & DistilGPT2 teacher, RTX PRO 6000 (M) & 342\,$\mu$s & 49.0\,mJ & 7.03\,mJ & 0.747 \\
     & DistilGPT2 circuit, Alveo U55C (E) & 0.97--1.75\,$\mu$s & 112--263\,$\mu$J & 16.9--39.7\,$\mu$J & 0.677 \\
     & DistilGPT2 circuit, Kria KV260 (E) & 2.14--3.97\,$\mu$s & 16.0--59.5\,$\mu$J & 2.7--9.6\,$\mu$J & 0.677 \\
    \midrule
    \multirow{8}{*}{SST-2} & BERT teacher, RTX PRO 6000 (M) & 461\,$\mu$s & 61.5\,mJ & 6.27\,mJ & 0.864 \\
     & BiBERT, Alveo U55C (E, bound) & $\ge$\,1.59--2.86\,$\mu$s & $\ge$\,183--429\,$\mu$J & $\ge$\,183--429\,$\mu$J & 0.798 \\
     & BiBERT, Kria KV260 (E, bound) & $\ge$\,122.9\,$\mu$s & $\ge$\,922--1,843\,$\mu$J & $\ge$\,142--511\,$\mu$J & 0.798 \\
     & BERT circuit, Alveo U55C (E) & 0.98--1.76\,$\mu$s & 112--264\,$\mu$J & 17.0--39.8\,$\mu$J & 0.788 \\
     & BERT circuit, Kria KV260 (E) & 2.19--4.06\,$\mu$s & 16.4--60.8\,$\mu$J & 2.7--9.9\,$\mu$J & 0.788 \\
     & DistilGPT2 teacher, RTX PRO 6000 (M) & 349\,$\mu$s & 49.6\,mJ & 6.98\,mJ & 0.883 \\
     & DistilGPT2 circuit, Alveo U55C (E) & 0.97--1.75\,$\mu$s & 112--262\,$\mu$J & 16.9--39.6\,$\mu$J & 0.777 \\
     & DistilGPT2 circuit, Kria KV260 (E) & 2.05--3.82\,$\mu$s & 15.4--57.3\,$\mu$J & 2.5--9.2\,$\mu$J & 0.777 \\
    \midrule
    \multirow{8}{*}{QQP} & BERT teacher, RTX PRO 6000 (M) & 459\,$\mu$s & 63.1\,mJ & 6.69\,mJ & 0.816 \\
     & BiBERT, Alveo U55C (E, bound) & $\ge$\,1.91--3.43\,$\mu$s & $\ge$\,219--515\,$\mu$J & $\ge$\,219--515\,$\mu$J & 0.725 \\
     & BiBERT, Kria KV260 (E, bound) & $\ge$\,122.9\,$\mu$s & $\ge$\,922--1,843\,$\mu$J & $\ge$\,170--612\,$\mu$J & 0.725 \\
     & BERT circuit, Alveo U55C (E) & 1.12--2.02\,$\mu$s & 129--303\,$\mu$J & 19.5--45.8\,$\mu$J & 0.705 \\
     & BERT circuit, Kria KV260 (E) & 2.91--5.36\,$\mu$s & 21.8--80.3\,$\mu$J & 3.8--13.6\,$\mu$J & 0.705 \\
     & DistilGPT2 teacher, RTX PRO 6000 (M) & 354\,$\mu$s & 52.0\,mJ & 8.24\,mJ & 0.801 \\
     & DistilGPT2 circuit, Alveo U55C (E) & 1.08--1.94\,$\mu$s & 124--291\,$\mu$J & 18.8--44.1\,$\mu$J & 0.707 \\
     & DistilGPT2 circuit, Kria KV260 (E) & 2.62--4.83\,$\mu$s & 19.6--72.4\,$\mu$J & 3.4--12.1\,$\mu$J & 0.707 \\
    \midrule
    \multirow{8}{*}{MRPC} & BERT teacher, RTX PRO 6000 (M) & 481\,$\mu$s & 72.1\,mJ & 11.55\,mJ & 0.809 \\
     & BiBERT, Alveo U55C (E, bound) & $\ge$\,3.28--5.90\,$\mu$s & $\ge$\,377--885\,$\mu$J & $\ge$\,377--885\,$\mu$J & 0.694 \\
     & BiBERT, Kria KV260 (E, bound) & $\ge$\,122.9\,$\mu$s & $\ge$\,922--1,843\,$\mu$J & $\ge$\,292--1,053\,$\mu$J & 0.694 \\
     & BERT circuit, Alveo U55C (E) & 1.75--3.14\,$\mu$s & 201--471\,$\mu$J & 30.6--71.8\,$\mu$J & 0.672 \\
     & BERT circuit, Kria KV260 (E) & 6.21--11.30\,$\mu$s & 46.6--169.4\,$\mu$J & 8.5--30.6\,$\mu$J & 0.672 \\
     & DistilGPT2 teacher, RTX PRO 6000 (M) & 357\,$\mu$s & 56.0\,mJ & 15.47\,mJ & 0.765 \\
     & DistilGPT2 circuit, Alveo U55C (E) & 1.83--3.30\,$\mu$s & 211--495\,$\mu$J & 33.1--76.8\,$\mu$J & 0.671 \\
     & DistilGPT2 circuit, Kria KV260 (E) & 8.88--16.11\,$\mu$s & 66.6--241.6\,$\mu$J & 12.5--45.0\,$\mu$J & 0.671 \\
    \midrule
    \multirow{8}{*}{CoLA} & BERT teacher, RTX PRO 6000 (M) & 445\,$\mu$s & 57.4\,mJ & 3.36\,mJ & 0.757 \\
     & BiBERT, Alveo U55C (E, bound) & $\ge$\,0.74--1.33\,$\mu$s & $\ge$\,85--200\,$\mu$J & $\ge$\,85--200\,$\mu$J & 0.662 \\
     & BiBERT, Kria KV260 (E, bound) & $\ge$\,122.9\,$\mu$s & $\ge$\,922--1,843\,$\mu$J & $\ge$\,66--238\,$\mu$J & 0.662 \\
     & BERT circuit, Alveo U55C (E) & 0.59--1.05\,$\mu$s & 67--158\,$\mu$J & 10.1--23.6\,$\mu$J & 0.671 \\
     & BERT circuit, Kria KV260 (E) & 0.90--1.74\,$\mu$s & 6.7--26.1\,$\mu$J & 1.0--3.5\,$\mu$J & 0.671 \\
     & DistilGPT2 teacher, RTX PRO 6000 (M) & 328\,$\mu$s & 44.8\,mJ & 3.10\,mJ & 0.742 \\
     & DistilGPT2 circuit, Alveo U55C (E) & 0.56--1.00\,$\mu$s & 64--150\,$\mu$J & 9.8--22.4\,$\mu$J & 0.670 \\
     & DistilGPT2 circuit, Kria KV260 (E) & 0.98--1.89\,$\mu$s & 7.4--28.3\,$\mu$J & 1.1--3.9\,$\mu$J & 0.670 \\
    \bottomrule
  \end{tabular}
\end{table}

\begin{table}[H]
  \caption{\RuleSel\ circuits per task, as \cref{tab:hw-tasks} for \orderSel: latency at batch 1 and energy per example in throughput mode on the Alveo U55C, latency and single-stream energy on the Kria KV260, and test gold accuracy (\cref{tab:main}). Estimated (E), mean over five seeds.}
  \label{tab:hw-rule-tasks}
  \centering
  \small
  \setlength{\tabcolsep}{4pt}
  \begin{tabular}{@{}llrrrrc@{}}
    \toprule
    & & \multicolumn{2}{c}{Alveo U55C} & \multicolumn{2}{c}{Kria KV260} & \\
    \cmidrule(lr){3-4}\cmidrule(lr){5-6}
    Task & Network & latency & energy, throughput & latency & energy, batch 1 & Gold \\
    \midrule
    \multirow{2}{*}{TweetEval-emotion} & BERT, 6 blocks & 0.64--1.16\,$\mu$s & 10.6--24.8\,$\mu$J & 1.99--3.71\,$\mu$s & 14.9--55.6\,$\mu$J & 0.618 \\
     & DistilGPT2 & 0.64--1.15\,$\mu$s & 10.5--24.6\,$\mu$J & 1.43--2.69\,$\mu$s & 10.7--40.3\,$\mu$J & 0.564 \\
    \midrule
    \multirow{2}{*}{SST-2} & BERT, 6 blocks & 0.64--1.15\,$\mu$s & 10.5--24.7\,$\mu$J & 0.93--1.80\,$\mu$s & 7.0--27.0\,$\mu$J & 0.747 \\
     & DistilGPT2 & 0.64--1.15\,$\mu$s & 10.5--24.6\,$\mu$J & 0.89--1.72\,$\mu$s & 6.7--25.8\,$\mu$J & 0.699 \\
    \midrule
    \multirow{2}{*}{QQP} & BERT, 6 blocks & 0.72--1.29\,$\mu$s & 11.8--27.7\,$\mu$J & 1.12--2.14\,$\mu$s & 8.4--32.1\,$\mu$J & 0.656 \\
     & DistilGPT2 & 0.70--1.25\,$\mu$s & 11.4--26.8\,$\mu$J & 1.02--1.95\,$\mu$s & 7.6--29.3\,$\mu$J & 0.659 \\
    \midrule
    \multirow{2}{*}{MRPC} & BERT, 6 blocks & 1.05--1.90\,$\mu$s & 17.3--40.7\,$\mu$J & 2.18--4.04\,$\mu$s & 16.3--60.6\,$\mu$J & 0.680 \\
     & DistilGPT2 & 1.04--1.88\,$\mu$s & 17.2--40.3\,$\mu$J & 1.82--3.40\,$\mu$s & 13.7--51.0\,$\mu$J & 0.656 \\
    \midrule
    \multirow{2}{*}{CoLA} & BERT, 6 blocks & 0.43--0.78\,$\mu$s & 7.1--16.6\,$\mu$J & 0.56--1.14\,$\mu$s & 4.2--17.0\,$\mu$J & 0.686 \\
     & DistilGPT2 & 0.41--0.74\,$\mu$s & 6.8--16.0\,$\mu$J & 0.54--1.10\,$\mu$s & 4.1--16.5\,$\mu$J & 0.687 \\
    \bottomrule
  \end{tabular}
\end{table}

\begin{table}[H]
  \caption{Components of every circuit per task (E, mean over five seeds): the design chosen for one Alveo U55C (units per block, lookup tables, registers, 36-kbit block RAMs and UltraRAMs) and the largest design that fits a Kria KV260 (units per block, lookup tables and their share of the device). Where the seeds differ, both unit counts are given. No design uses a DSP block or a multiplier.}
  \label{tab:hw-comp-tasks}
  \centering
  \small
  \setlength{\tabcolsep}{3.5pt}
  \begin{tabular}{@{}lllrrrrrrr@{}}
    \toprule
    & & & \multicolumn{5}{c}{Alveo U55C} & \multicolumn{2}{c}{Kria KV260} \\
    \cmidrule(lr){4-8}\cmidrule(l){9-10}
    Task & Network & Selection & units & LUTs & registers & BRAM36 & URAM & units & LUTs \\
    \midrule
    \multirow{4}{*}{TweetEval-emotion} & BERT, 6 blocks & order & 32 & 784,502 & 821,743 & 22 & 98 & 2 & 76,556 (65\%) \\
     & BERT, 6 blocks & rule & 63 & 671,810 & 664,489 & 12 & 98 & 6/7 & 75,743 (65\%) \\
     & DistilGPT2 & order & 63 & 515,581 & 374,309 & 15 & 162 & 7/8 & 76,867 (66\%) \\
     & DistilGPT2 & rule & 63 & 422,678 & 405,056 & 12 & 162 & 10/11 & 78,504 (67\%) \\
    \midrule
    \multirow{4}{*}{SST-2} & BERT, 6 blocks & order & 63 & 511,356 & 374,178 & 18 & 98 & 7 & 76,090 (65\%) \\
     & BERT, 6 blocks & rule & 63 & 168,604 & 114,459 & 12 & 98 & 26/27/29 & 80,628 (69\%) \\
     & DistilGPT2 & order & 63 & 517,704 & 378,197 & 14 & 162 & 7/8 & 79,815 (68\%) \\
     & DistilGPT2 & rule & 63 & 154,581 & 103,639 & 12 & 162 & 31/32/33 & 81,125 (69\%) \\
    \midrule
    \multirow{4}{*}{QQP} & BERT, 6 blocks & order & 63 & 549,776 & 419,570 & 18 & 98 & 6/7 & 78,313 (67\%) \\
     & BERT, 6 blocks & rule & 63 & 179,904 & 131,032 & 12 & 98 & 25/26/28/29 & 80,548 (69\%) \\
     & DistilGPT2 & order & 63 & 519,169 & 382,180 & 16 & 162 & 7 & 76,209 (65\%) \\
     & DistilGPT2 & rule & 63 & 163,448 & 112,539 & 12 & 162 & 29/30/31 & 80,575 (69\%) \\
    \midrule
    \multirow{4}{*}{MRPC} & BERT, 6 blocks & order & 63 & 521,814 & 393,536 & 18 & 98 & 7 & 77,913 (67\%) \\
     & BERT, 6 blocks & rule & 63 & 183,019 & 136,139 & 12 & 98 & 24/26/27/28 & 80,445 (69\%) \\
     & DistilGPT2 & order & 32/63 & 769,342 & 726,136 & 13 & 162 & 4/5 & 76,551 (65\%) \\
     & DistilGPT2 & rule & 63 & 149,133 & 97,554 & 12 & 162 & 31/32/33/34/35 & 81,040 (69\%) \\
    \midrule
    \multirow{4}{*}{CoLA} & BERT, 6 blocks & order & 63 & 522,678 & 386,925 & 17 & 98 & 7 & 76,604 (65\%) \\
     & BERT, 6 blocks & rule & 63 & 173,309 & 122,085 & 12 & 98 & 26/28/31 & 80,544 (69\%) \\
     & DistilGPT2 & order & 32/63 & 717,508 & 672,554 & 15 & 162 & 4 & 75,462 (64\%) \\
     & DistilGPT2 & rule & 63 & 138,653 & 85,358 & 12 & 162 & 33/34/36/38 & 80,632 (69\%) \\
    \bottomrule
  \end{tabular}
\end{table}

\begin{table}[H]
  \caption{Teachers on an RTX 4090 with CUDA graphs (M), per task: latency and energy per example at batch 1, and energy per example and throughput at batch 256.}
  \label{tab:gpu4090-tasks}
  \centering
  \small
  \begin{tabular}{@{}llrrrr@{}}
    \toprule
    Task & Network & latency, batch 1 & energy, batch 1 & energy, batch 256 & examples/s, batch 256 \\
    \midrule
    \multirow{2}{*}{TweetEval-emotion} & BERT, 6 blocks & 526\,$\mu$s & 99.2\,mJ & 14.19\,mJ & 30,709 \\
     & DistilGPT2 & 403\,$\mu$s & 86.8\,mJ & 13.47\,mJ & 32,123 \\
    \midrule
    \multirow{2}{*}{SST-2} & BERT, 6 blocks & 530\,$\mu$s & 94.2\,mJ & 13.88\,mJ & 31,322 \\
     & DistilGPT2 & 406\,$\mu$s & 85.0\,mJ & 13.38\,mJ & 32,393 \\
    \midrule
    \multirow{2}{*}{QQP} & BERT, 6 blocks & 538\,$\mu$s & 105.7\,mJ & 17.08\,mJ & 25,722 \\
     & DistilGPT2 & 412\,$\mu$s & 81.8\,mJ & 15.87\,mJ & 26,776 \\
    \midrule
    \multirow{2}{*}{MRPC} & BERT, 6 blocks & 555\,$\mu$s & 119.9\,mJ & 29.37\,mJ & 14,939 \\
     & DistilGPT2 & 433\,$\mu$s & 102.8\,mJ & 29.75\,mJ & 14,454 \\
    \midrule
    \multirow{2}{*}{CoLA} & BERT, 6 blocks & 485\,$\mu$s & 86.4\,mJ & 5.92\,mJ & 73,378 \\
     & DistilGPT2 & 388\,$\mu$s & 78.0\,mJ & 5.12\,mJ & 85,162 \\
    \bottomrule
  \end{tabular}
\end{table}

\begin{table}[H]
  \caption{Hardware of the main-line circuits over five seeds (mean $\pm$ s.d.\ where the seeds differ). Gold accuracy of the hardware on test, lookup tables (LUTs) of one unit per block after synthesis, the design chosen for one Alveo U55C (units per block, LUT utilisation, latency of the logic per tweet) and the largest design that fits a Kria KV260 (units per block, LUT utilisation, latency of the logic), all estimated (E) as in \cref{tab:hw}.}
  \label{tab:hw-seeds}
  \centering
  \small
  \setlength{\tabcolsep}{2.5pt}
  \begin{tabular}{@{}llcccccc@{}}
    \toprule
    & & & LUTs, one unit & \multicolumn{2}{c}{Alveo U55C} & \multicolumn{2}{c}{Kria KV260} \\
    \cmidrule(lr){5-6}\cmidrule(l){7-8}
    Model & Selection & Gold & per block & units (LUTs) & latency, $\mu$s & units (LUTs) & latency, $\mu$s \\
    \midrule
    \multirow{2}{*}{DistilGPT2} & order & 0.677\,$\pm$\,0.008 & 28,174\,$\pm$\,507 & 63 (39.6\,$\pm$\,0.9\%) & 0.97--1.75 & 7.2\,$\pm$\,0.4 (65.6\,$\pm$\,2.2\%) & 2.04--3.67 \\
     & rule & 0.564\,$\pm$\,0.014 & 17,124\,$\pm$\,548 & 63 (32.4\,$\pm$\,1.2\%) & 0.64--1.15 & 10.4\,$\pm$\,0.5 (67.0\,$\pm$\,0.9\%) & 1.33--2.39 \\
    \midrule
    \multirow{2}{*}{BERT-base, 6 blocks} & order & 0.711\,$\pm$\,0.012 & 52,958\,$\pm$\,1,037 & 32 (60.2\,$\pm$\,1.3\%) & 1.08--1.94 & 2 (65.4\,$\pm$\,1.3\%) & 5.08--9.14 \\
     & rule & 0.618\,$\pm$\,0.050 & 21,232\,$\pm$\,1,377 & 63 (51.5\,$\pm$\,1.7\%) & 0.64--1.16 & 6.2\,$\pm$\,0.4 (64.6\,$\pm$\,2.7\%) & 1.89--3.41 \\
    \bottomrule
  \end{tabular}
\end{table}

\begin{table}[H]
  \caption{Status of every quantity in the hardware comparison.}
  \label{tab:hw-status}
  \centering
  \small
  \begin{tabular}{@{}p{0.2\textwidth}p{0.11\textwidth}p{0.63\textwidth}@{}}
    \toprule
    Quantity & Status & Basis \\
    \midrule
    Accuracy and agreement & measured & the software circuit on the test split (\cref{tab:main,tab:ablation}) \\
    Bit-exact export & measured & integer evaluator of the exported description against the software circuit on every token state and class of the test tweets; synthesised netlists simulated on real inputs, 0 mismatches \\
    GPU throughput, latency, board power & measured & teacher in eager PyTorch, FP16 and FP32, tweets grouped by length without padding, throughput at batch 256, latency at batch 1, board power sampled about every 50\,ms over windows of at least 20\,s, on an RTX 3090, an RTX 5090 and an RTX PRO 6000; also with CUDA graphs in FP16 (DistilGPT2 on the RTX 5090, the 6-block BERT on the RTX PRO 6000), one captured graph per input shape, predictions equal to the eager ones on all test tweets \\
    B200 and B300 throughput & estimated & anchored on the measured RTX PRO 6000, from launch-bound (same throughput) to compute-bound (the smaller of the dense FP16 and memory-bandwidth ratios) \\
    FPGA area & estimated & Yosys and ABC synthesis to 6-input lookup tables and UltraScale+ resource counts, no place-and-route; a design fits a device if it uses at most 70\% of its lookup tables and registers; control logic and host interface assumed \\
    Latency of the logic & estimated & cycle count of the pipelined design at the assumed clocks, mean over the lengths of the test tweets \\
    Clock & estimated & 250, 300 and 450\,MHz, bracketed by the platform's default kernel clock, the device limits of memories and clock buffers, and published lookup-table netlists on the same family \citep{umuroglu2020logicnets} \\
    Dynamic power & estimated & per-LUT, per-register and per-block-RAM energies derived from published Vivado power breakdowns of 16\,nm designs \citep{plagwitz2023spike}, toggle rates of 10--30\% \\
    Board and host power & sourced, estimated & AMD typical and maximum board power, CPU, memory and chipset figures and power-supply efficiency; energy per tweet bracketed from typical to maximum power \\
    Tokenisation & measured, estimated & Rust tokenizers measured on a laptop processor, scaled to desktop processors and to Cortex-A53 cores by CoreMark scores, with an assumed factor of 0.5--1 for the smaller caches of the A53 \\
    Prices & sourced & launch, list or reseller prices of GPUs, host parts and boards; per-GPU shares of 8-GPU systems for B200 and B300; storage, network switches and cabling not costed \\
    \bottomrule
  \end{tabular}
\end{table}

\begin{table}[H]
  \caption{Throughput and energy per tweet for the \orderSel\ circuit of DistilGPT2. M: measured, E: estimated. (a) Card only, without tokenisation. GPU throughput and energy in FP16 at batch 256. FPGA ranges span clocks of 250--450\,MHz and typical to maximum board power. (b) Equal whole-system cost, end to end with tokenisation, against an x86 desktop with an RTX 5090.}
  \label{tab:hw-tput}
  \centering
  \small
  \begin{tabular}{@{}llrr@{}}
    \toprule
    & System (price) & Tweets/s & Energy per tweet \\
    \midrule
    (a) card & RTX 3090 (\$1,499), teacher in PyTorch (M) & 14,845 & 22.6\,mJ \\
    & RTX 5090 (\$1,999), teacher in PyTorch (M) & 45,226 & 12.1\,mJ \\
    & RTX 5090 (\$1,999), teacher with CUDA graphs (M) & 51,166 & -- \\
    & Alveo U55C (\$4,747), circuit logic (E) & 3.8--6.8\,M & 16.9--39.7\,$\mu$J \\
    \midrule
    (b) system & x86 desktop with RTX 5090 (\$3,260), teacher (M; host power E) & 45,226 & 16.4--18.1\,mJ \\
    & two tokenising desktops and one Kria KV260 board (\$3,084), circuit (E) & 355k--434k & 1.06--1.56\,mJ \\
    & ratio, circuit system to GPU system & 7.9--9.6$\times$ & 10.5--16.9$\times$ less \\
    \bottomrule
  \end{tabular}
\end{table}

\begin{table}[H]
  \caption{Equal-cost comparison over four classes of GPU, end to end with tokenisation. For each GPU system, the circuit system is the best of the configurations searched at the same whole-system budget: x86 desktops (PCs) that tokenise and feed shared FPGA boards. Throughput in tweets/s, energy per tweet of the whole system. GPU side: M measured, E estimated; circuit side: estimated. \OrderSel\ circuits, the same for all five seeds. Gold accuracy: teachers 0.747 (DistilGPT2) and 0.800 (6-block BERT), circuits in \cref{tab:main}.}
  \label{tab:gpus}
  \centering
  \footnotesize
  \setlength{\tabcolsep}{2.5pt}
  \begin{tabular}{@{}lrrlrrrr@{}}
    \toprule
    GPU (system cost) & \multicolumn{2}{c}{GPU system} & Circuit system (cost) & \multicolumn{2}{c}{Circuit system} & \multicolumn{2}{c}{Ratio} \\
    \cmidrule(lr){2-3}\cmidrule(lr){5-6}\cmidrule(l){7-8}
    & tweets/s & mJ/tweet & & tweets/s & mJ/tweet & tweets/s & energy \\
    \midrule
    \multicolumn{8}{@{}l}{\emph{DistilGPT2}} \\
    RTX 5090 (\$3.26k) & 45.2k M & 16.4--18.1 & 2 PCs + 1 KV260 (\$3.08k) & 355k--434k & 1.06--1.56 & 7.9--9.6$\times$ & 10.5--16.9$\times$ \\
    RTX PRO 6000 (\$25.6k) & 38.9k M & 12.6--14.4 & 17 PCs + 3 KV260 (\$24.7k) & 3.02--3.69\,M & 1.05--1.50 & 78--95$\times$ & 8.4--13.7$\times$ \\
    B200 (\$63.9k) & 38.9k--174k E & 10.3--46 & 44 PCs + 6 KV260 (\$63.5k) & 7.82--9.56\,M & 1.05--1.50 & 45--246$\times$ & 6.9--43.7$\times$ \\
    B300 (\$77.0k) & 38.9k--174k E & 10.4--48.5 & 53 PCs + 7 KV260 (\$76.4k) & 9.42--11.51\,M & 1.05--1.50 & 54--296$\times$ & 7.0--46.1$\times$ \\
    \midrule
    \multicolumn{8}{@{}l}{\emph{BERT-base, 6 blocks}} \\
    RTX 5090 (\$3.26k) & 58.1k M & 13.0--14.4 & 2 PCs + 1 KV260 (\$3.08k) & 274k--335k & 1.26--2.25 & 4.7--5.8$\times$ & 5.8--11.4$\times$ \\
    RTX PRO 6000 (\$25.6k) & 85.0k M & 8.0--8.8 & 17 PCs + 5 KV260 (\$25.3k) & 2.33--2.84\,M & 1.25--2.20 & 27--34$\times$ & 3.6--7.1$\times$ \\
    B200 (\$63.9k) & 85.0k--380k E & 4.7--21.0 & 43 PCs + 12 KV260 (\$63.7k) & 5.89--7.20\,M & 1.25--2.20 & 16--85$\times$ & 2.1--16.8$\times$ \\
    B300 (\$77.0k) & 85.0k--380k E & 4.8--22.2 & 52 PCs + 14 KV260 (\$76.9k) & 7.12--8.70\,M & 1.25--2.19 & 19--102$\times$ & 2.2--17.8$\times$ \\
    \bottomrule
  \end{tabular}
\end{table}

\paragraph{Reading the GPU classes.} The circuit side scales linearly with the budget, at a constant throughput per dollar, and it is limited by tokenisation on commodity desktop processors. Measured GPU throughput changes little between these GPUs for this workload: the RTX PRO 6000 reaches 0.86 times the throughput of the RTX 5090 on DistilGPT2 and 1.46 times on the 6-block BERT, at 7.9 times the system cost. The B200 and B300 estimates add at most 4.5 times at 20--24 times the cost. The equal-cost ratio therefore grows with the price of the GPU, mainly because the dearer GPUs are poor value for the inference of short tweets, and the RTX 5090, the most cost-effective GPU here, is the comparison we report in the main text. The B200 and B300 budgets come from vendor listings and press reports.

\paragraph{Scope of the throughput comparison.} The GPU baselines of the equal-cost comparison run the teacher in eager PyTorch FP16, the way it was trained and evaluated. With CUDA graphs, the measured batch-1 latency drops from 2.21 to 0.89\,ms for DistilGPT2 on the RTX 5090 and from 1.47 to 0.46\,ms for the 6-block BERT on the RTX PRO 6000, and the batch-256 throughput is 51,166 and 68,730 tweets/s (eager: 45,226 and 85,026). A faster GPU system also becomes bound by tokenisation, so that the equal-cost systems move closer together. The comparison of circuit logic and GPU card remains the largest effect.

\paragraph{Boards that tokenise on their own processors.} Standalone ARM+FPGA boards that tokenise on their four Cortex-A53 cores, without an x86 host, are limited by that tokenisation at about 1.8--3.5k tweets/s per board, hundreds of times below their own fabric. At the cost of an x86 desktop with an RTX 3090 (\$2,705), nine Kria KV260 boards (\$2,466) reach an estimated 15.8k--31.6k tweets/s end to end for DistilGPT2, 1.1--2.1 times the measured 14,845 tweets/s of the GPU system, with 4.3--8.5\,mJ instead of 34--39\,mJ per tweet. For the 6-block BERT, whose \orderSel\ circuit fits a KV260 with two units per block, nine boards reach 12.2k--24.4k tweets/s, 0.59--1.18 times the measured 20,678 tweets/s of the GPU system, with 0.9--10.1 times less energy per tweet.

\paragraph{The 6-block BERT at the card level.} The estimated circuit logic of the 6-block BERT on one Alveo U55C processes 3.32--5.97\,M tweets/s with 19.3--45.2\,$\mu$J per tweet at board power, 39--70 times the measured throughput of the fastest GPU for this teacher (RTX PRO 6000, 85,026 tweets/s at batch 256) with 127--299 times less energy per tweet, and its latency is 756--1,361 times shorter than the 1.47\,ms of the RTX PRO 6000 at batch 1 in eager PyTorch and 238--429 times shorter than its 0.46\,ms with CUDA graphs. On a KV260, its logic takes 5.1--9.1\,$\mu$s per tweet, and end to end with tokenisation on the board's A53 cores 1.48--2.97\,ms.

\paragraph{The rule-selection circuits.} The \ruleSel\ circuits are exported and verified in the same way, with 0 mismatches on 236,055 checked rows of the synthesised netlists of each. Their selection needs no score tables and no row maximum, and their logic takes an estimated 0.64--1.15\,$\mu$s (DistilGPT2) and 0.64--1.16\,$\mu$s (6-block BERT) per tweet on a U55C and 1.33--2.39 and 1.89--3.41\,$\mu$s on a KV260, mean over five seeds (\cref{tab:hw-seeds}).

\subsection{State-machine merging before export}
\label{app:merge}

Before export, every circuit is minimised exactly as a state machine, in two steps. First, the states that the circuit's own tables can never produce at an interface are removed, together with the literals and rule terms that need them. Second, the states that nothing above can tell apart are merged, from the readout down, by partition refinement as in the minimisation of sequential machines \citep{huffman1954synthesis}: two states of an interface stay in one class only if they have the same query and key codes, no remaining count literal refers to them, their vote rows at the readout are equal, and their next states fall into one class. Lexical codes that nothing above can tell apart are merged likewise. The minimised circuit computes the same function. On every token and interface of the fit set and of the validation and test splits, its states fall into the classes of the original states and its readout agrees with the original one, with 0 mismatches, and its synthesised netlists agree with it on all checked rows. The minimisation runs once, offline, in minutes per circuit.

\begin{table}[H]
  \caption{State-machine merging before export (seed 0, teacher-correct fit set). States: sum over the interfaces, of which the unproducible ones are removed and the indistinguishable ones merged. Bits: the tables, rules and votes of the exported description without the input table, which merging leaves unchanged (selection rules in their exported form, not as the one-bit tables of \cref{tab:size}). LUTs and registers: one unit per block after synthesis, and the LUTs of the design chosen for the Alveo U55C. Gold accuracy and agreement with the teacher on test.}
  \label{tab:merge}
  \centering
  \small
  \setlength{\tabcolsep}{4pt}
  \begin{tabular}{@{}lrrrrrrrrcc@{}}
    \toprule
    & \multicolumn{3}{c}{States} & Lexical & & \multicolumn{2}{c}{One unit per block} & LUTs & & \\
    \cmidrule(lr){2-4}\cmidrule(lr){7-8}
    Merging & total & removed & merged & codes & Bits & LUTs & registers & (U55C) & Gold & Agr. \\
    \midrule
    \multicolumn{11}{@{}l}{\emph{DistilGPT2, \orderSel}} \\
    \quad without & 480 & -- & -- & 257 & 385,046 & 29,172 & 17,740 & 529,078 & 0.683 & 0.710 \\
    \quad with & 346 & 134 & 0 & 256 & 370,511 & 29,048 & 17,622 & 529,264 & 0.683 & 0.710 \\
    \multicolumn{11}{@{}l}{\emph{DistilGPT2, \ruleSel}} \\
    \quad without & 480 & -- & -- & 257 & 574,077 & 17,618 & 12,569 & 438,164 & 0.548 & 0.564 \\
    \quad with & 423 & 57 & 0 & 256 & 567,928 & 17,535 & 12,491 & 434,981 & 0.548 & 0.564 \\
    \multicolumn{11}{@{}l}{\emph{BERT-base, 6 blocks, \orderSel}} \\
    \quad without & 480 & -- & -- & 257 & 619,127 & 52,340 & 43,134 & 768,998 & 0.723 & 0.768 \\
    \quad with & 436 & 44 & 0 & 257 & 614,535 & 52,302 & 43,090 & 769,146 & 0.723 & 0.768 \\
    \multicolumn{11}{@{}l}{\emph{BERT-base, 6 blocks, \ruleSel}} \\
    \quad without & 480 & -- & -- & 257 & 705,220 & 19,448 & 15,254 & 660,218 & 0.543 & 0.549 \\
    \quad with & 421 & 59 & 0 & 257 & 699,165 & 19,277 & 15,101 & 653,289 & 0.543 & 0.549 \\
    \bottomrule
  \end{tabular}
\end{table}

\Cref{tab:merge} compares the four circuits without and with merging (seed 0). Almost the whole reduction is the removal of states that the circuit cannot produce: no state is merged proper and one lexical code is merged in each of the two DistilGPT2 circuits, so the states that the conversion fits are already distinguishable by the circuit above (\cref{sec:interfaces}). Merging lowers the bits of the tables, rules and votes by 0.7--3.8\% and the lookup tables and registers of one unit by 0.1--1.0\%. It leaves the chosen number of units per card and board, the logic depth, the latency and the throughput unchanged, and it changes the lookup tables of the chosen U55C design and the dynamic energy per tweet by at most 1.0\%. Gold accuracy and agreement are identical.

\section{Partial hardening and circuit size}
\label{app:size}

\begin{figure}[H]
  \centering
  \begin{tikzpicture}
    \begin{groupplot}[
      group style={group size=2 by 2, horizontal sep=16mm, vertical sep=17mm},
      width=0.45\textwidth, height=4.6cm,
      xlabel={blocks replaced by circuits},
      grid=major, grid style={black!10}, axis line style={black!45}, tick style={black!45},
      tick label style={font=\footnotesize}, label style={font=\footnotesize}, title style={font=\footnotesize},
      yticklabel style={/pgf/number format/fixed, /pgf/number format/precision=2},
      legend style={font=\footnotesize, draw=none, fill=white, fill opacity=0.85, text opacity=1, at={(0.03,0.03)}, anchor=south west},
      legend cell align=left,
    ]
      \nextgroupplot[title={DistilGPT2: test gold accuracy}, xmin=-0.3, xmax=6.3, xtick={0,...,6}, ymin=0.6, ymax=0.84]
      \addplot[black!55, dashed, thick, domain=-0.3:6.3, samples=2] {0.788};
      \addplot[seriesorder, line width=1pt, mark=*, mark size=1.8pt] coordinates {(0,0.788) (1,0.789) (2,0.783) (3,0.767) (4,0.744) (5,0.714) (6,0.700)};
      \addplot[seriesrule, line width=1pt, mark=square*, mark size=1.6pt] coordinates {(0,0.788) (1,0.788) (2,0.776) (3,0.748) (4,0.724) (5,0.689) (6,0.653)};
      \legend{teacher, \orderSel, \ruleSel}
      \nextgroupplot[title={DistilGPT2: agreement with the teacher on test}, xmin=-0.3, xmax=6.3, xtick={0,...,6}, ymin=0.6, ymax=1.02]
      \addplot[seriesorder, line width=1pt, mark=*, mark size=1.8pt] coordinates {(0,1) (1,0.926) (2,0.902) (3,0.862) (4,0.827) (5,0.771) (6,0.737)};
      \addplot[seriesrule, line width=1pt, mark=square*, mark size=1.6pt] coordinates {(0,1) (1,0.922) (2,0.869) (3,0.824) (4,0.788) (5,0.728) (6,0.674)};
      \legend{\orderSel, \ruleSel}
      \nextgroupplot[title={BERT-base, first 6 blocks: test gold accuracy}, xmin=-0.3, xmax=6.3, xtick={0,...,6}, ymin=0.6, ymax=0.84]
      \addplot[black!55, dashed, thick, domain=-0.3:6.3, samples=2] {0.809};
      \addplot[seriesorder, line width=1pt, mark=*, mark size=1.8pt] coordinates {(0,0.809) (1,0.785) (2,0.770) (3,0.750) (4,0.721) (5,0.716) (6,0.709)};
      \addplot[seriesrule, line width=1pt, mark=square*, mark size=1.6pt] coordinates {(0,0.809) (1,0.778) (2,0.749) (3,0.725) (4,0.688) (5,0.686) (6,0.678)};
      \legend{teacher, \orderSel, \ruleSel}
      \nextgroupplot[title={BERT-base, first 6 blocks: agreement with the teacher on test}, xmin=-0.3, xmax=6.3, xtick={0,...,6}, ymin=0.6, ymax=1.02]
      \addplot[seriesorder, line width=1pt, mark=*, mark size=1.8pt] coordinates {(0,1) (1,0.916) (2,0.878) (3,0.837) (4,0.788) (5,0.776) (6,0.761)};
      \addplot[seriesrule, line width=1pt, mark=square*, mark size=1.6pt] coordinates {(0,1) (1,0.903) (2,0.843) (3,0.796) (4,0.743) (5,0.738) (6,0.717)};
      \legend{\orderSel, \ruleSel}
    \end{groupplot}
  \end{tikzpicture}
  \caption{Every stage of the conversion is a complete classifier. Test gold accuracy and agreement with the teacher on test of the hybrid after the top $k$ blocks have been replaced by circuits, with the teacher's own blocks below, fitted on the teacher-correct training examples, mean over the five tasks and five seeds, both forms of selection: DistilGPT2 (top) and BERT-base truncated to its first 6 blocks (bottom). $k=0$ is the FP8 teacher (dashed: its mean gold accuracy), and the last point is the complete circuit.}
  \label{fig:curve}
\end{figure}
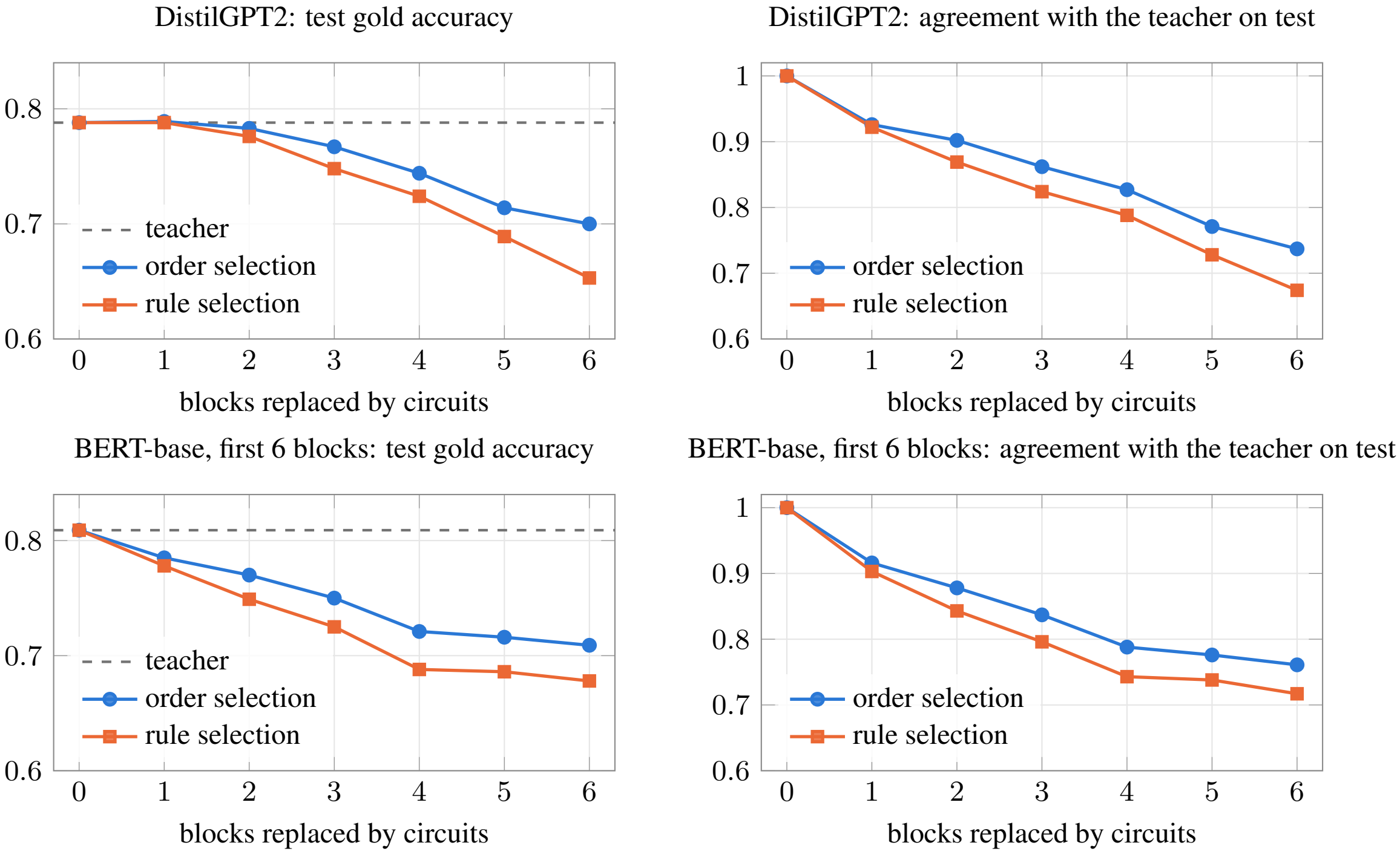

With \orderSel, DistilGPT2 with one circuit block keeps the gold accuracy of its teacher on average (0.789 against 0.788) and exceeds it on TweetEval-emotion (0.759 against 0.747). The agreement with the teacher goes from 0.926 and 0.916 after the first block to 0.737 and 0.761 for the complete \orderSel\ circuits of DistilGPT2 and the 6-block BERT (\cref{tab:stages-tasks,tab:stages-tasks-rule}).

\begin{table}[H]
  \caption{Stages per task (\orderSel, mean over five seeds): test gold accuracy and agreement with the teacher on test after the top $k$ blocks have become circuits, with the teacher's own blocks below. $k=0$ is the FP8 teacher and $k=6$ the complete circuit of \cref{tab:main}.}
  \label{tab:stages-tasks}
  \centering
  \small
  \begin{tabular}{@{}lllccccccc@{}}
    \toprule
    & & & \multicolumn{7}{c}{blocks converted, $k$} \\
    \cmidrule(l){4-10}
    Model & Task & & 0 & 1 & 2 & 3 & 4 & 5 & 6 \\
    \midrule
    \multirow{10}{*}{DistilGPT2} & \multirow{2}{*}{TweetEval-emotion} & gold & .747 & .759 & .745 & .731 & .717 & .691 & .677 \\
     & & agr. & 1 & .899 & .861 & .823 & .786 & .746 & .711 \\
     & \multirow{2}{*}{SST-2} & gold & .883 & .882 & .880 & .859 & .829 & .787 & .777 \\
     & & agr. & 1 & .972 & .954 & .906 & .863 & .823 & .802 \\
     & \multirow{2}{*}{QQP} & gold & .801 & .805 & .801 & .791 & .774 & .739 & .707 \\
     & & agr. & 1 & .948 & .914 & .890 & .857 & .801 & .742 \\
     & \multirow{2}{*}{MRPC} & gold & .765 & .763 & .757 & .756 & .740 & .693 & .671 \\
     & & agr. & 1 & .900 & .902 & .879 & .864 & .749 & .689 \\
     & \multirow{2}{*}{CoLA} & gold & .742 & .734 & .730 & .695 & .662 & .658 & .670 \\
     & & agr. & 1 & .912 & .877 & .814 & .762 & .736 & .739 \\
    \midrule
    \multirow{10}{*}{BERT, 6 blocks} & \multirow{2}{*}{TweetEval-emotion} & gold & .800 & .782 & .760 & .744 & .721 & .729 & .711 \\
     & & agr. & 1 & .913 & .869 & .834 & .787 & .793 & .752 \\
     & \multirow{2}{*}{SST-2} & gold & .864 & .841 & .830 & .814 & .798 & .796 & .788 \\
     & & agr. & 1 & .935 & .906 & .873 & .847 & .840 & .827 \\
     & \multirow{2}{*}{QQP} & gold & .816 & .806 & .797 & .779 & .747 & .711 & .705 \\
     & & agr. & 1 & .922 & .888 & .850 & .801 & .742 & .722 \\
     & \multirow{2}{*}{MRPC} & gold & .809 & .754 & .739 & .717 & .691 & .676 & .672 \\
     & & agr. & 1 & .892 & .849 & .815 & .769 & .753 & .753 \\
     & \multirow{2}{*}{CoLA} & gold & .757 & .742 & .724 & .697 & .650 & .668 & .671 \\
     & & agr. & 1 & .920 & .880 & .814 & .737 & .750 & .750 \\
    \bottomrule
  \end{tabular}
\end{table}

\begin{table}[H]
  \caption{Stages per task with \ruleSel, as \cref{tab:stages-tasks} for \orderSel\ (mean over five seeds).}
  \label{tab:stages-tasks-rule}
  \centering
  \small
  \begin{tabular}{@{}lllccccccc@{}}
    \toprule
    & & & \multicolumn{7}{c}{blocks converted, $k$} \\
    \cmidrule(l){4-10}
    Model & Task & & 0 & 1 & 2 & 3 & 4 & 5 & 6 \\
    \midrule
    \multirow{10}{*}{DistilGPT2} & \multirow{2}{*}{TweetEval-emotion} & gold & .747 & .753 & .705 & .675 & .657 & .610 & .564 \\
     & & agr. & 1 & .891 & .796 & .738 & .708 & .637 & .582 \\
     & \multirow{2}{*}{SST-2} & gold & .883 & .883 & .881 & .846 & .806 & .757 & .699 \\
     & & agr. & 1 & .974 & .951 & .890 & .839 & .791 & .712 \\
     & \multirow{2}{*}{QQP} & gold & .801 & .806 & .795 & .788 & .759 & .708 & .659 \\
     & & agr. & 1 & .946 & .904 & .884 & .837 & .764 & .677 \\
     & \multirow{2}{*}{MRPC} & gold & .765 & .776 & .783 & .756 & .732 & .695 & .656 \\
     & & agr. & 1 & .922 & .868 & .840 & .823 & .712 & .670 \\
     & \multirow{2}{*}{CoLA} & gold & .742 & .724 & .714 & .675 & .667 & .675 & .687 \\
     & & agr. & 1 & .880 & .825 & .766 & .735 & .736 & .728 \\
    \midrule
    \multirow{10}{*}{BERT, 6 blocks} & \multirow{2}{*}{TweetEval-emotion} & gold & .800 & .755 & .744 & .718 & .662 & .655 & .618 \\
     & & agr. & 1 & .886 & .841 & .792 & .714 & .706 & .633 \\
     & \multirow{2}{*}{SST-2} & gold & .864 & .842 & .820 & .799 & .784 & .762 & .747 \\
     & & agr. & 1 & .922 & .883 & .849 & .829 & .802 & .769 \\
     & \multirow{2}{*}{QQP} & gold & .816 & .804 & .757 & .747 & .697 & .674 & .656 \\
     & & agr. & 1 & .914 & .804 & .794 & .726 & .695 & .655 \\
     & \multirow{2}{*}{MRPC} & gold & .809 & .760 & .727 & .671 & .629 & .660 & .680 \\
     & & agr. & 1 & .899 & .841 & .764 & .706 & .739 & .772 \\
     & \multirow{2}{*}{CoLA} & gold & .757 & .732 & .698 & .688 & .666 & .677 & .686 \\
     & & agr. & 1 & .895 & .845 & .781 & .738 & .749 & .754 \\
    \bottomrule
  \end{tabular}
\end{table}

\begin{table}[H]
  \caption{Size of the complete circuits by part (teacher-correct fit set, seed 0), \orderSel\ / \ruleSel. Literals count the two-level logic of the minimised tables and rules, and Mbit counts table storage, with the input table over the whole vocabulary at every position. The vote readout reads 2,262 / 276 predicates for DistilGPT2 and 3,960 / 240 for the 6-block BERT. With \ruleSel\ the score tables are not part of the circuit, and the selection rules (757 and 760 literals as logic) are stored as compiled one-bit tables. Teachers: every parameter at 8 bits, 81,916,416 parameters for DistilGPT2 (with the added padding token and the four-class head) and 66,958,084 for the 6-block BERT-base.}
  \label{tab:size}
  \centering
  \small
  \begin{tabular}{@{}lcccc@{}}
    \toprule
    & \multicolumn{2}{c}{DistilGPT2} & \multicolumn{2}{c}{BERT-base, 6 blocks} \\
    \cmidrule(lr){2-3}\cmidrule(l){4-5}
    Part & Literals & Mbit & Literals & Mbit \\
    \midrule
    Input table $\Lambda$ & -- & 47.49 / 47.49 & -- & 28.84 / 28.84 \\
    Code maps & 117,382 / 117,211 & 0.12 / 0.12 & 118,372 / 118,045 & 0.12 / 0.12 \\
    Score tables & 109,283 / -- & 0.17 / -- & 110,550 / -- & 0.17 / -- \\
    Selection rules & -- / 757 & -- / 20.05 & -- / 760 & -- / 20.05 \\
    Transition rules & 1,436 / 6,123 & -- & 16,069 / 7,718 & -- \\
    Vote readout & -- & 0.07 / 0.01 & -- & 0.13 / 0.01 \\
    \midrule
    Complete circuit & 228,101 / 124,091 & 47.86 / 67.68 & 244,991 / 126,523 & 29.26 / 49.03 \\
    Teacher (FP8) & -- & 655.3 & -- & 535.7 \\
    \bottomrule
  \end{tabular}
\end{table}

\begin{table}[H]
  \caption{Size of every system per task: whole vocabulary and active vocabulary (the word pieces that occur in the task) in Mbit, and bits read per example (\cref{app:hw}), by the same rule for every system. Circuits: mean over five seeds.}
  \label{tab:size-tasks}
  \centering
  \small
  \begin{tabular}{@{}llrrr@{}}
    \toprule
    Task & System & whole & active & read per example \\
    \midrule
    \multirow{7}{*}{TweetEval-emotion} & BERT teacher (FP8) & 535.7 & 423.0 & 345.3\,Mbit \\
     & BiBERT & 75.0 & 60.6 & 45.1\,Mbit \\
     & BERT circuit, order selection & 29.3 & 11.9 & 0.38\,kbit \\
     & BERT circuit, rule selection & 49.0 & 31.7 & 0.38\,kbit \\
     & DistilGPT2 teacher (FP8) & 655.3 & 453.0 & 340.6\,Mbit \\
     & DistilGPT2 circuit, order selection & 47.8 & 16.7 & 0.38\,kbit \\
     & DistilGPT2 circuit, rule selection & 67.7 & 36.5 & 0.38\,kbit \\
    \midrule
    \multirow{7}{*}{SST-2} & BERT teacher (FP8) & 535.7 & 414.5 & 345.3\,Mbit \\
     & BiBERT & 75.0 & 59.5 & 45.1\,Mbit \\
     & BERT circuit, order selection & 29.2 & 10.6 & 0.38\,kbit \\
     & BERT circuit, rule selection & 49.0 & 30.4 & 0.38\,kbit \\
     & DistilGPT2 teacher (FP8) & 655.3 & 424.5 & 340.5\,Mbit \\
     & DistilGPT2 circuit, order selection & 47.8 & 12.3 & 0.38\,kbit \\
     & DistilGPT2 circuit, rule selection & 67.7 & 32.2 & 0.38\,kbit \\
    \midrule
    \multirow{7}{*}{QQP} & BERT teacher (FP8) & 535.7 & 448.0 & 345.4\,Mbit \\
     & BiBERT & 75.0 & 63.8 & 45.2\,Mbit \\
     & BERT circuit, order selection & 29.2 & 15.7 & 0.45\,kbit \\
     & BERT circuit, rule selection & 49.0 & 35.5 & 0.45\,kbit \\
     & DistilGPT2 teacher (FP8) & 655.3 & 493.6 & 340.6\,Mbit \\
     & DistilGPT2 circuit, order selection & 47.8 & 23.0 & 0.43\,kbit \\
     & DistilGPT2 circuit, rule selection & 67.7 & 42.8 & 0.43\,kbit \\
    \midrule
    \multirow{7}{*}{MRPC} & BERT teacher (FP8) & 535.7 & 421.6 & 345.6\,Mbit \\
     & BiBERT & 75.0 & 60.4 & 45.5\,Mbit \\
     & BERT circuit, order selection & 29.2 & 11.7 & 0.78\,kbit \\
     & BERT circuit, rule selection & 49.0 & 31.5 & 0.78\,kbit \\
     & DistilGPT2 teacher (FP8) & 655.3 & 437.8 & 340.9\,Mbit \\
     & DistilGPT2 circuit, order selection & 47.8 & 14.3 & 0.77\,kbit \\
     & DistilGPT2 circuit, rule selection & 67.7 & 34.2 & 0.77\,kbit \\
    \midrule
    \multirow{7}{*}{CoLA} & BERT teacher (FP8) & 535.7 & 384.2 & 345.1\,Mbit \\
     & BiBERT & 75.0 & 55.7 & 44.9\,Mbit \\
     & BERT circuit, order selection & 29.2 & 5.9 & 0.18\,kbit \\
     & BERT circuit, rule selection & 49.0 & 25.7 & 0.18\,kbit \\
     & DistilGPT2 teacher (FP8) & 655.3 & 387.8 & 340.4\,Mbit \\
     & DistilGPT2 circuit, order selection & 47.8 & 6.7 & 0.16\,kbit \\
     & DistilGPT2 circuit, rule selection & 67.7 & 26.5 & 0.16\,kbit \\
    \bottomrule
  \end{tabular}
\end{table}

\Cref{tab:size-tasks} gives the sizes of every system per task. The input table dominates every total. The converted blocks of the \orderSel\ circuits take 0.37\,Mbit of tables and 228,101 literals of logic for DistilGPT2, and 0.42\,Mbit and 244,991 literals for the 6-block BERT, whose complete circuit is 18.3 times smaller than its FP8 teacher.

\section{Proofs}
\label{app:theory}

Let $z_0=\Lambda(x)$ and $z_{m+1}=\Gamma_m(z_m)$, and let the circuit predict $\rho(z_L)$, where $z_L$ denotes the bank of the readout token read by the votes. Fix interface targets $\tau_0=\Lambda(x),\tau_1(x),\dots,\tau_L(x)$, and write $u\eqR{\mathcal{R}}v$ when $u$ and $v$ agree on everything in $\mathcal{R}$ at every token. Here $\mathcal{R}_{m+1}$ is what $\Gamma_{m+1}$ reads, so that merged states are equal under $\eqR{\mathcal{R}_{m+1}}$. The \emph{contract error} of block $m$ is $\epsilon_m=\Pr_x[\Gamma_m(\tau_m)\neqR{\mathcal{R}_{m+1}}\tau_{m+1}]$, and that of the readout is $\epsilon_L=\Pr_x[\rho(\tau_L)\ne y]$.

\begin{proposition}[Contract errors add]
\label{prop:additive}
$\Pr_x[\rho(z_L)\ne y(x)]\le\sum_{m=0}^{L}\epsilon_m$.
\end{proposition}

\begin{proof}[Proof of Proposition~\ref{prop:additive}]
On the event that no contract fails, $z_m\eqR{\mathcal{R}_m}\tau_m$ holds for every $m$ by induction. It holds for $m=0$. If it holds for $m$, then $\Gamma_m(z_m)=\Gamma_m(\tau_m)$ because $\Gamma_m$ reads only $\mathcal{R}_m$, and $\Gamma_m(\tau_m)\eqR{\mathcal{R}_{m+1}}\tau_{m+1}$. Hence $\rho(z_L)=\rho(\tau_L)=y$. A union bound over the failure events completes the proof.
\end{proof}

\begin{proposition}[Occam bound for rule sets]
\label{prop:occam}
Fix a prefix-free code for rule sets, independently of the sample, and let $\ell(g)$ be the length in bits of the description of rule set $g$. Let $N$ examples $(x,y(x))$ be drawn i.i.d., with input literals given by a map fixed independently of them. With probability at least $1-\eta$, every rule set $g$ satisfies
\begin{equation*}
  \mathrm{err}(g)\le\widehat{\mathrm{err}}(g)+\sqrt{\frac{\ell(g)\ln 2+\ln(1/\eta)}{2N}}.
\end{equation*}
\end{proposition}

\begin{proof}[Proof of Proposition~\ref{prop:occam}]
Apply Hoeffding's inequality to each $g$ at confidence $\eta\,2^{-\ell(g)}$ and take a union bound. Kraft's inequality gives $\sum_g 2^{-\ell(g)}\le 1$ for a prefix-free code \citep{blumer1987occam,shalevshwartz2014understanding}.
\end{proof}

\end{document}